\documentclass[lettersize,journal]{IEEEtran}
\usepackage{amsmath,amsfonts}
\usepackage{algorithm}
\usepackage{array}
\usepackage[caption=false,font=normalsize,labelfont=sf,textfont=sf]{subfig}
\usepackage{textcomp}
\usepackage{stfloats}
\usepackage{url}
\usepackage{verbatim}
\usepackage{graphicx}
\usepackage{cite}
\usepackage{multirow}
\usepackage{xcolor}

\usepackage{longtable}
\usepackage{booktabs}
\usepackage{tabularx} 
\usepackage{utfsym}
\usepackage{colortbl}
\usepackage{color}

\usepackage{pifont}

\usepackage{soul}
\usepackage{amssymb}
\usepackage{bbding}
\usepackage{subcaption}

\usepackage{newfloat}
\usepackage{listings}
\usepackage{epsfig}
\usepackage{caption}
\usepackage{setspace}
\usepackage{algpseudocode}
\usepackage{tabu}
\usepackage{hyperref}

\begin{document}

\title{Towards an Effective Action-Region Tracking Framework for Fine-grained Video Action Recognition}

\author{Baoli Sun, Yihan Wang, Xinzhu Ma, Zhihui Wang,  Kun Lu, Zhiyong Wang
\thanks{This work was supported by National Key R\&D Program of China under Grant 2025ZD0121800, and Australian Research Council (ARC) Discovery Project under Grant DP210102674. (Corresponding author: Zhihui Wang, Email: zhwang@dlut.edu.cn.)}
\thanks{B. Sun, Y. Wang, Z. Wang, K. Lu are with DUT-RU International School of Information Science \& Engineering, Dalian University of Technology, China.}
\thanks{X. Ma is with Beihang University, China.}
\thanks{Z. Wang is with The University of Sydney, Australia.}
}

\markboth{Journal of \LaTeX\ Class Files,~Vol.~14, No.~8, August~2021}%
{Shell \MakeLowercase{\textit{et al.}}: A Sample Article Using IEEEtran.cls for IEEE Journals}

\maketitle

\begin{abstract}
Fine-grained action recognition (FGAR) aims to identify subtle and distinctive differences among fine-grained action categories.
    However, current recognition methods often capture coarse-grained motion patterns but struggle to identify subtle details in local regions evolving over time.
    In this work, we introduce the Action-Region Tracking (ART) framework, a novel solution leveraging a query-response mechanism to discover and track the dynamics of distinctive local details, enabling distinguishing similar actions effectively.
    Specifically, we propose a region-specific semantic activation module that employs discriminative and text-constrained semantics serve as queries to capture the most action-related  region responses in each video frame, facilitating interaction among spatial and temporal dimensions with corresponding video features. 
    The captured region responses are then organized into action tracklets, which characterize the region-based action dynamics by linking related responses across different video frames in a coherent sequence.
    The text-constrained queries are designed to expressly encode nuanced semantic representations derived from the textual descriptions of action labels, as extracted by the language branches within Visual Language Models (VLMs).
    To optimize generated action tracklets, we design a multi-level tracklet contrastive constraint among multiple region responses at spatial and temporal levels, which can effectively distinguish individual region responses in each video frame (spatial level) and establish the correlation of similar region responses between adjacent video frames (temporal level).
    Additionally, we implement a task-specific fine-tuning mechanism to refine textual semantics during training. This ensures that the semantic representations encoded by VLMs are not only preserved but also optimized for specific task preferences.
    Comprehensive experiments on several widely used action recognition benchmarks, \textit{i.e.}, FineGym, Diving48, NTURGB-D, Kinetics, and Something-Something, clearly demonstrate the superiority to previous state-of-the-art baselines.
\end{abstract}


\section{Introduction}
\label{sec:intro}

\IEEEPARstart{F}{ine-grained}  action recognition (FGAR) is aimed at distinguishing subtle and discriminative differences within fine-grained action categories, aligning more closely with the increasing complexity of human activities in the real world. 
This emerging area has driven the evolution of action recognition tasks toward a finer granularity, attracting  significant attention within the research community due to its potential to enhance various visual analytics applications, including intelligent surveillance \cite{DBLP:journals/tnn/Atto24a}, social scene understanding \cite{DBLP:journals/tnn/FuWDXDLL24}, and sports video analysis \cite{DBLP:journals/tnn/GaoYWD24}.

Recently, video action recognition research has advanced significantly, driven by the emergence of powerful model architectures \cite{DBLP:conf/eccv/WangXW0LTG16} \cite{DBLP:conf/iccv/QiuYM17} \cite{DBLP:conf/iccv/JiangWGWY19} \cite{DBLP:conf/cvpr/ShaoZDL20} \cite{DBLP:conf/cvpr/YangXSDZ20} \cite{DBLP:journals/tnn/AlfaslyCJLX24} \cite{DBLP:conf/cvpr/Zhang0Z21} and the availability of large-scale datasets \cite{DBLP:conf/cvpr/CarreiraZ17} \cite{DBLP:journals/corr/abs-2010-10864} \cite{DBLP:conf/iccv/GoyalKMMWKHFYMH17} \cite{DBLP:conf/cvpr/ShaoZDL20a}. The inherent challenge of video action recognition lies in addressing how to effective encode the spatio-temporal representations. Existing methods can be broadly categorized into two groups: spatio-temporal feature modeling methods (\textit{e.g.}, TSM \cite{DBLP:conf/iccv/LinGH19}, ATM \cite{DBLP:conf/iccv/WuSS0XO23} and Uniformer \cite{DBLP:journals/pami/LiWZGSLLQ23}) and long-temporal modeling methods (\textit{e.g.},SlowFast \cite{DBLP:conf/iccv/Feichtenhofer0M19}, TDN \cite{DBLP:conf/cvpr/0002TJW21} and  MViT \cite{DBLP:conf/cvpr/YanXALZ0S22}).
For example, TSM \cite{DBLP:conf/iccv/LinGH19} enhances motion pattern representation by integrating spatio-temporal features with feature-level motion encoding, allowing for a more dynamic capture of motion. Uniformer \cite{DBLP:journals/pami/LiWZGSLLQ23} proposes to effectively combines 3D convolution with spatio-temporal self-attention mechanisms within a streamlined transformer architecture, thereby achieving a preferable balance between efficiency and effectivenes. TDN \cite{DBLP:conf/cvpr/0002TJW21} addresses multi-scale temporal dynamics by differentiating and integrating temporal features across short-term and long-term intervals. MViT \cite{DBLP:conf/cvpr/YanXALZ0S22} introduces a multi-view transformer architecture for video recognition that employs distinct encoders for processing various temporal segments, with lateral connections to fuse information across views.

\begin{figure*}[!h]
	\centering
	\centerline{\includegraphics[width=1\linewidth]{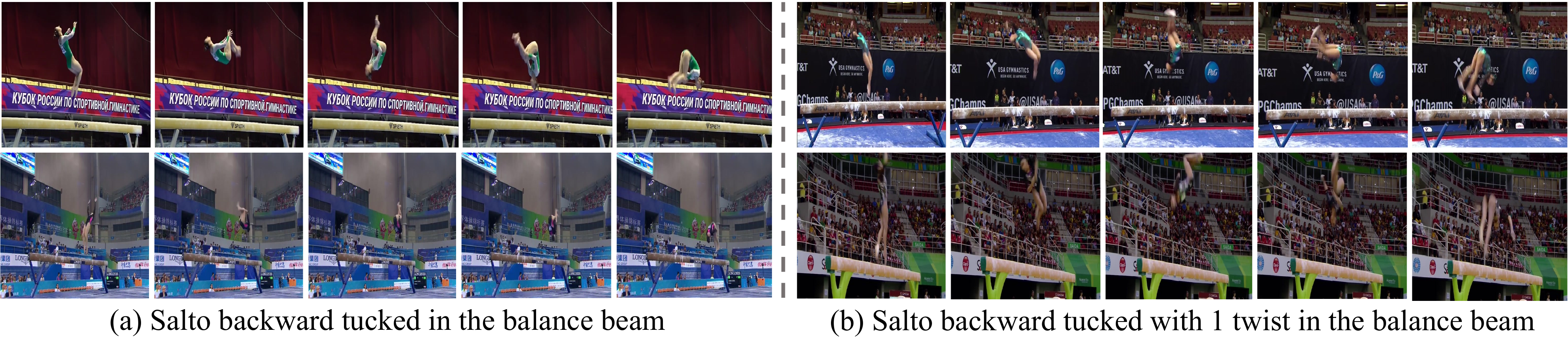}}
	\caption{Examples of \textit{“salto backward tucked”} and its variant with 1 twist reveal the fine-grained nature of action recognition, characterized by large intra-class variation and subtle inter-class differences. Accurate discrimination relies more on capturing temporal dynamics of local movements than on appearance or context.}
	\label{fig:motiv0}
\end{figure*}

In previous general action recognition  methods, visual appearance cues of objects and background often  play a crucial role, sometimes even more significant than the action itself. For example, in distinguishing between actions like "\textit{basketball}" and "\textit{gymnastics}", the scene information alone can easily allow for differentiation.
FGAC is significantly more challenging than general action recognition, primarily due to subtle inter-class differences. These differences are often more pronounced in the temporal dimension, where motion information is dominant over visual appearance.
Specifically, key motion differences are observed in targeted regions of the human body, including the pose adjustments, the extent of movement range, and the dynamic interaction among limbs.
For instance, Figure \ref{fig:motiv0} illustrates the balance beam event in gymnastics, where fine-grained actions such as "\textit{salto backward tucked}" and "\textit{salto backward tucked with 1 twist}" are subtly differentiated by the gymnast's execution of a twist during the backward flip.
Consequently, the recognition models should have a robust capability for discovering and representing distinct local details and their dynamics, which is essential for achieving accurate fine-grained action understanding.

Substantial  efforts have been dedicated to the exploration of crucial region localization and interaction in vision recognition tasks \cite{DBLP:conf/eccv/FayyazKJSJSPG22} \cite{DBLP:conf/nips/RaoZLLZH21} \cite{DBLP:conf/cvpr/Zeng0000O022}  \cite{DBLP:journals/tomccap/FengXMZ24} \cite{DBLP:conf/mm/SunYWLW23} \cite{DBLP:conf/iccv/ChenTSWW23}. For instance,  Fayyaz et al. \cite{DBLP:conf/eccv/FayyazKJSJSPG22} developed an adaptive token sampler for vision transformers that scores and selectively samples crucial tokens, improving efficiency without compromising performance.
STMixer \cite{DBLP:journals/pami/WuCGWW24}  introduced a query-based adaptive feature sampling module to flexibly extract discriminative spatiotemporal features, yielding an efficient end-to-end action detector.
TempMe \cite{DBLP:conf/iclr/ShenHHZZLBD25} proposed a parameter-efficient and computation-efficient framework for text-video retrieval by addressing temporal redundancy across video frames. TempMe leveraged a progressive multi-granularity temporal token merging strategy to significantly reduce trainable parameters, token count, and model complexity. 
ALIN \cite{DBLP:conf/mm/SunYWLW23} dynamically identified the most informative tokens at a coarse granularity and subsequently refined these located tokens to a finer granularity, enabling the exploration of valuable fine-grained spatio-temporal interactions.
These approaches aim to diminish the number of tokens processed during various feature extraction phases, thereby significantly reducing computational costs.
However, this selective filtering process often results in a set of discontinuous tokens across both space and time, disrupting the continuity of action regions.
Specifically, if tokens identified as crucial in one frame are missing or only partially captured in subsequent frames, it complicates the accurate modeling of action sequences, leading to potential inaccuracies in recognition tasks.
A straightforward solution involves explicitly locating and tracking discriminative, action-specific local regions as an auxiliary task. However, the resource-intensive nature of video data makes training end-to-end models that manage both localization and tracking particularly challenging, especially in the absence of detailed pixel-level or region-level annotations.

\begin{figure*}[!h]
	\centering
	\centerline{\includegraphics[width=1\linewidth]{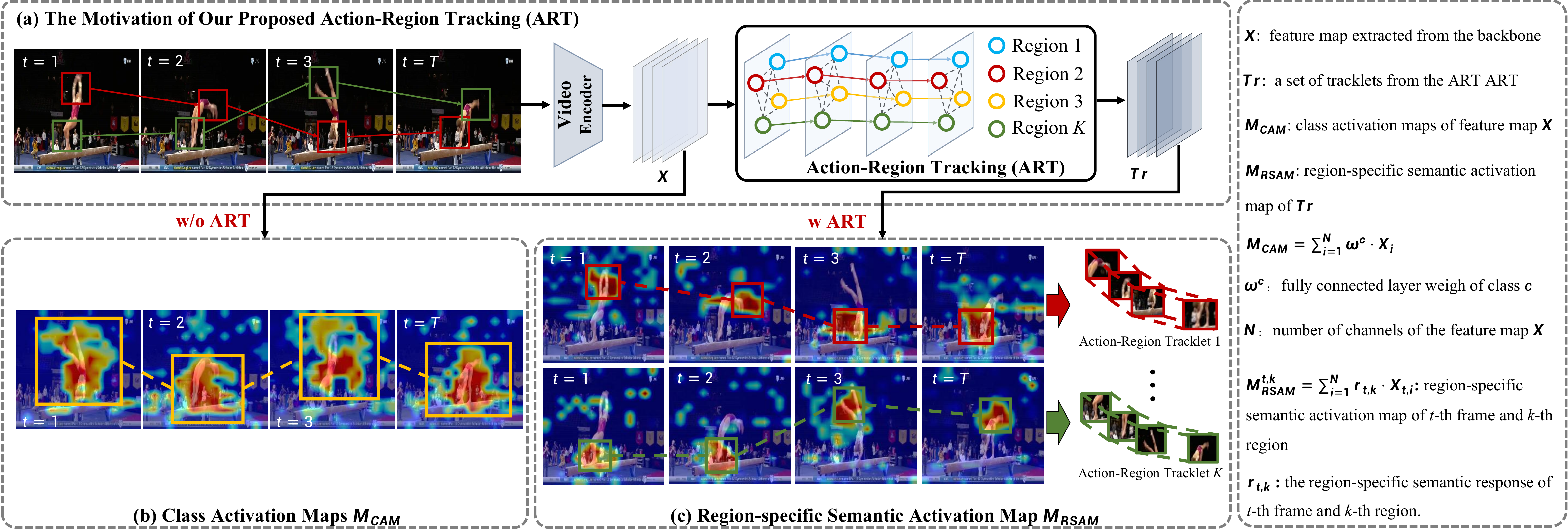}}
	\caption{(a) The visualization illustrates the motivation behind our Action-Region Tracking (ART) framework. Our ART aims to identifies and tracks discriminative action regions across multiple local areas that evolve over time. Here, $\textbf{\emph{X}}$ represents semantic features extracted from the backbone network, while $\textbf{\emph{Tr}}$ denotes features processed by ART.  (b) and (c) give the class activation maps and the response regions contributing to action prediction without and with ART, respectively. We can see that the backbone network tends to concentrate on easily distinguishable regions, often overlooking the dynamics of local details. In contrast, Our ART framework focuses on discriminative action regions over time}
	\label{fig:motiv}
\end{figure*}

These limitations motivate us to design a streamlined and effective framework, which implicitly identifies and tracks distinctive local details along the temporal dimension in a self-supervised manner, where only video-level labels are available.
Building on the success of query-based mechanisms in transformers for object detection (\textit{e.g.}, EDTR \cite{DBLP:conf/eccv/CarionMSUKZ20} and UP-DETR \cite{DBLP:conf/cvpr/DaiCLC21}) and instance segmentation (\textit{e.g.}, VisTR \cite{DBLP:conf/cvpr/WangXWSCSX21} and EfficientVIS \cite{Wu_2022_CVPR}), we adapt this approach to track action-specific regions in video sequences.
Rather than relying on implicit spatio-temporal encoding, explicitly discovering and
organizing local action regions into tracklets facilitates better reasoning over motion
dynamics. These tracklets provide localized, structured motion representations
that not only enhance recognition accuracy but also improve the interpretability of
model predictions. 
To this end, we propose a novel Action-Region Tracking framework, termed ART, as shown in Figure \ref{fig:net}.
Initially, ART employs a standard video encoder to sequentially extract region-level features from the input video.
Secondly, ART introduces a region-specific semantic activation module that employs discriminative and text-constrained semantics serve as queries to capture the most action-related  region responses in each video frame, facilitating interaction among spatial and temporal dimensions with corresponding video features. 
The text-constrained queries are designed to expressly encode nuanced semantic representations derived from the textual descriptions of action labels, as extracted by the language branches within Visual Language Models (VLMs) \cite{DBLP:conf/icml/RadfordKHRGASAM21}. Notably, these semantic representations are stored in memory, allowing for efficient retrieval without real-time processing by VLMs during inference.
Subsequently, the region-specific semantic responses are organized into a group of action tracklets, in which each tracklet links related responses over time, facilitating detailed regional action tracking.
In Figure \ref{fig:motiv} (c), we illustrate how individual region responses are grouped into action-region tracklets that follow consistent spatial-temporal dynamics. These tracklets are aligned to meaningful sub-actions (e.g., takeoff, mid-air twist), offering an intuitive depiction of how ART disentangles motion cues.
In contrast, the class activation maps (as shown in Figure \ref{fig:motiv} (b)) of backbone network tend to highlight broad and coarse regions, such as the entire body, without exhibiting spatial selectivity toward meaningful sub-parts.
This process disentangles the overall spatio-temporal representation into distinct  tracklets corresponding to different body parts, thus sharpening the semantic distinctions and enhancing the characterization of fine-grained action differences.

However, in the absence of region-level annotations, accurately constraining these region queries to capture specific action details becomes difficult.
To address this, we introduce a Multi-level Tracklet Contrastive Loss that operates on region-aware semantic responses at spatial, temporal, and tracklet levels. It captures diverse semantic responses in each frame and models high correlations among similar response regions across adjacent frames.
This loss function serves as a robust self-supervised constraint that: (1) distinguishes individual region responses within each frame at the spatial level, (2) establishes correlations among region responses across adjacent frames at the temporal level, and (3) promotes the generation of diverse tracklets at the tracklet level.

Given the minimal variations among category names and the fact that the quality of action-region localization is determined by their semantic representations, this situation can result in a text semantic space with insufficient discriminative power. This lack of specificity may introduce ambiguity in the final recognition process.
To mitigate this issue, we consider the adaptability of semantic representations to each individual video instance, tailoring them more closely to the unique characteristics of each video.
Initially, we introduce a set of learnable prompts designed to integrate semantic representations into the region-specific semantic activation module. This integration allows for the nuanced capture of individual variations within each video.
Furthermore, we implement a task-specific fine-tuning mechanism inspired by the Exponential Moving Average (EMA) approach. This mechanism refines the textual semantics during training, preserving the core semantic representations encoded by VLMs while optimizing them for task-specific preferences.

The proposed ART demonstrates robust performance in fine-grained action recognition, achieving 91.2\% top-1 accuracy on FineGym99 and 84.4\% top-1 accuracy on FineGym288. These results significantly surpass the current state-of-the-art.
This represents a clear advancement over the previous state-of-the-art model, MDCN \cite{DBLP:conf/mm/SunYYWLW22}, improving by margins of +1.3\% and +1.0\% on FineGym99 and FineGym288, respectively.
Additionally, ART achieves these results with reduced computational complexity compared to full transformer-based methods such as ViViT \cite{DBLP:journals/corr/abs-2103-15691} and MViT \cite{DBLP:conf/iccv/0001XMLYMF21}.
The key contributions are summarized as follows:

$\bullet$
We propose a novel action-region tracking framework (ART) to discover and track distinctive details of local regions in a video frame to enrich the encoding of spatial and temporal contexts for better reasoning fine-grained actions.

$\bullet$
We devise a multi-level tracklet contrastive loss to guide ART to accurately capture action details in an effective self-supervised manner at spatial, temporal, and tracklet levels.

$\bullet$
We introduce a task-specific fine-tuning mechanism designed to enhance textual semantics, which retains the semantic representations encoded by VLMs and optimizes them to align with the specific requirements of the downstream task.

$\bullet$
The results of extensive experiments conducted on four widely used fine-grained and conventional action recognition datasets, including FineGym \cite{DBLP:conf/cvpr/ShaoZDL20a}, Diving48 \cite{DBLP:conf/eccv/LiLV18} NTURGB-D \cite{DBLP:journals/pami/LiuSPWDK20} and Kinetics \cite{DBLP:conf/cvpr/CarreiraZ17}, 
clearly demonstrate the superiority of our proposed method against the state-of-the-art.

\section{Related Work}
\label{related}

\subsection{Action Recognition}
Recognizing human behaviors is a fundamental  task in computer vision, involving the classification of observed actions based on video input. This process is pivotal for applications ranging from surveillance to interactive systems.
Typically, there are four types of deep-based models: two-stream model, 3D model, (2+1)D model, transformer-based model.
\textbf{(1) Two-stream model}: The two-stream architectures \cite{DBLP:conf/nips/SimonyanZ14} \cite{DBLP:conf/cvpr/FeichtenhoferPZ16} \cite{DBLP:conf/cvpr/FeichtenhoferPW17a} processes RGB frames and optical flow images through separate CNNs to extract features, which are then combined. This model excels in capturing motion but may lack temporal context in its analysis.
\textbf{(2) 3D model}: Since videos can be viewed as temporally dense sequences of image samples, extending 2D convolution operations of 2D CNN to 3D convolutions is the most intuitive method for spatio-temporal feature learning. 
For example, various 3D convolutional networks \cite{DBLP:conf/iccv/TranBFTP15} \cite{DBLP:conf/cvpr/CarreiraZ17}  \cite{DBLP:conf/iccv/Feichtenhofer0M19}
\cite{DBLP:conf/wacv/StroudRSDS20} have been proposed to directly learn spatio-temporal features from RGB frames.
C3D \cite{DBLP:conf/iccv/TranBFTP15} was the first work utilizing deep 3D CNNs for learning spatio-temporal features.
I3D \cite{DBLP:conf/cvpr/CarreiraZ17} introduced a new two-stream inflated 3D network that models spatial and motion features with a visual stream and a flow stream.
SlowFast \cite{DBLP:conf/iccv/Feichtenhofer0M19} involved a slow pathway and a fast pathway to capture spatial semantics at a low frame rate and motion information at a fast frame rate, respectively. 
\textbf{(3) (2+1)D model}: 3D networks generally suffer from heavy computational cost. 
To reduce computational cost, various methods \cite{DBLP:conf/iccv/QiuYM17}   \cite{DBLP:conf/cvpr/TranWTRLP18} \cite{DBLP:conf/iccv/JiangWGWY19} \cite{DBLP:conf/iccv/LinGH19} \cite{DBLP:conf/cvpr/LiJSZKW20} were proposed to decompose 3D convolutions into 2D spatial and 1D temporal filters.
STM \cite{DBLP:conf/iccv/JiangWGWY19} learns spatio-temporal and motion features from shared feature maps, while TEA \cite{DBLP:conf/cvpr/LiJSZKW20} leverages temporal differences with a hierarchical residual design to capture both short- and long-range motions.
\textbf{(4) Transformer-based model}: Inspired by the significant advancements achieved by Transformers in Natural Language Processing, the Vision Transformer (ViT) \cite{DBLP:conf/nips/VaswaniSPUJGKP17} was naturally developed from image recognition tasks to video recognition tasks \cite{DBLP:journals/corr/abs-2103-15691} \cite{DBLP:conf/icml/BertasiusWT21} \cite{DBLP:conf/cvpr/YanXALZ0S22} \cite{DBLP:journals/pami/LiWZGSLLQ23} \cite{DBLP:journals/tnn/ZhengACWR23} \cite{DBLP:journals/tnn/AlfaslyCJLX24} \cite{DBLP:juniformerv2iccv23}.
ViViT \cite{DBLP:journals/corr/abs-2103-15691} extended pure transformers to video classification.
TimeSformer \cite{DBLP:conf/icml/BertasiusWT21} tailored self-attention for spatio-temporal learning.
MViT \cite{DBLP:conf/cvpr/YanXALZ0S22} adopted multi-view encoders with lateral connections to fuse temporal information.

Despite progress, existing models struggle with videos differing only in subtle motions and temporal dynamics, often overlooking local nuances. To address this, our work enhances recognition of fine-grained motion and temporal details, advancing action recognition performance.

\subsection{Fine-grained Action Recognition}
Fine-grained action recognition, which focuses on distinguishing subtle inter-class differences among closely related action categories, has garnered increasing attention due to its complexity and applicability.
Compared to general action recognition, fine-grained action recognition is distinguished by several critical characteristics: (1) significant intra-class variation coupled with subtle inter-class differences, (2) the insufficiency of sparse sampling frames for representing actions, (3) the paramount importance of motion information over visual appearance, and (4) the necessity for precise temporal dynamics modeling.
As a foundation for more complex technologies, the pursuit of better datasets in the field of action understanding has never ceased.
Early, several datasets have been developed to facilitate research in this area, such as  Diving48 \cite{DBLP:conf/eccv/LiLV18}, Something-Something \cite{DBLP:conf/iccv/GoyalKMMWKHFYMH17}, NTU-RGBD \cite{DBLP:journals/pami/LiuSPWDK20}, each designed to explore nuanced human-object interactions.
For instance, Diving48 \cite{DBLP:conf/eccv/LiLV18} focused on the recognition and analysis of complex diving actions, classifying performances based on various actions and orientations like \textit{somersaults}, \textit{twists}, and \textit{handstands}.
Something-Something \cite{DBLP:conf/iccv/GoyalKMMWKHFYMH17} encompassed 147 classes of routine human-object interactions, capturing diverse activities \textit{ranging from moving objects down} to \textit{retrieving items from specific locations}.
Most recently, FineGym \cite{DBLP:conf/cvpr/ShaoZDL20a}, crafted from high-definition gymnasium videos, included intricate motion details exemplified by actions like \textit{vault-women: double salto backward tucked}. FineGym not only provided a new and large-scale benchmark but also had verified that current models were still inadequate in capturing the nuanced spatio-temporal semantics required for fine-grained recognition.
The focus of research is toward to the development of  effective approaches for fine-grained video understanding.
TQN \cite{DBLP:conf/cvpr/Zhang0Z21} leverages multi-attribute sub-labels from action texts for granular attribute learning with transformers. 
MDCN \cite{DBLP:conf/mm/SunYYWLW22} disentangles motion features to emphasize dominant dynamics.
CANet \cite{DBLP:journals/chinaf/ChengLGH23} employs class-specific attention and dictionary learning to decouple class-wise features.
PGVT \cite{DBLP:conf/wacv/0001LLL24} integrates pose priors and temporal attention for enhanced fine-grained action recognition.

Despite these advancements, existing models have not fully addressed the need for discriminative local motion representations that consider the dynamics of local action details. In this work, we propose to discover and track different action regions along the temporal dimension for more accurately characterizing fine-grained actions.

\subsection{Discriminative Region Localization}
The human brain employs a hierarchical and diverse scale of attention to process visual information, boosting its capacity to discern vital information from the environment while selectively ignoring the insignificant details. 
Leveraging the characteristics of serialized tokens in vision transformers and their capacity to capture long-range temporal dependencies, significant research efforts have been focused on pinpointing and examining crucial regions (tokens) and their interactions in the realm of image and video recognition tasks \cite{DBLP:conf/nips/RaoZLLZH21, DBLP:conf/iccv/WangX0FSLL0021,  DBLP:conf/eccv/FayyazKJSJSPG22, DBLP:conf/cvpr/Zeng0000O022, DBLP:journals/tnn/YaoWGPPTY24, DBLP:conf/mm/SunYWLW23, DBLP:conf/aaai/ChenLLSW0J23, DBLP:journals/corr/abs-2202-07800}. The patterns of discriminative region localization can broadly be categorized into three types: (1) \textbf{Attention based} \cite{DBLP:conf/nips/PatrickCAMMFVH21, DBLP:conf/cvpr/GritsenkoXD00LS24, DBLP:journals/pami/KorbanYA24, DBLP:journals/mva/YangWLP23}:
Motionformer \cite{DBLP:conf/nips/PatrickCAMMFVH21} employs trajectory attention to model long-range temporal dependencies via continuous token paths, while Korban et al. \cite{DBLP:journals/pami/KorbanYA24} propose a spatiotemporal transformer with semantic attention and motion-aware encoding to capture spatial–motion interactions and dynamic variations.
(2) \textbf{Token sampling} \cite{DBLP:conf/eccv/FayyazKJSJSPG22, DBLP:conf/nips/RaoZLLZH21, DBLP:conf/mm/SunYWLW23, DBLP:journals/corr/abs-2402-00033}:
retaining partially class-discriminative tokens within the sequence of tokens. 
DynamicViT \cite{DBLP:conf/nips/RaoZLLZH21} introduced a framework for dynamic token sparsification, aimed at progressively pruning redundant tokens. C2F-ALIN \cite{DBLP:conf/mm/SunYWLW23} dynamically identified the most informative tokens with coarse granularity and subsequently divided these tokens into finer granularity to facilitate detailed spatio-temporal interaction. 
(3) \textbf{Token fusion} \cite{DBLP:journals/corr/abs-2202-07800, DBLP:conf/iccv/WangX0FSLL0021, DBLP:conf/cvpr/Zeng0000O022}: aggregating token representations based on their semantics through the depth of a visual transformer. 
EViT \cite{DBLP:journals/corr/abs-2202-07800} enhances token selection by retaining attentive tokens and fusing inattentive ones via gradient back-propagation, while ToMe \cite{DBLP:conf/iclr/BolyaFDZFH23} accelerates transformers by merging similar tokens through a lightweight matching algorithm, combining pruning speed with improved accuracy.

In addition to the works that aim to accelerate the inference of convolutional neural networks, other works aim to improve the efficiency of transformer-based models.

\begin{figure*}[!t]
	\centering
	\centerline{\includegraphics[width=1\linewidth]{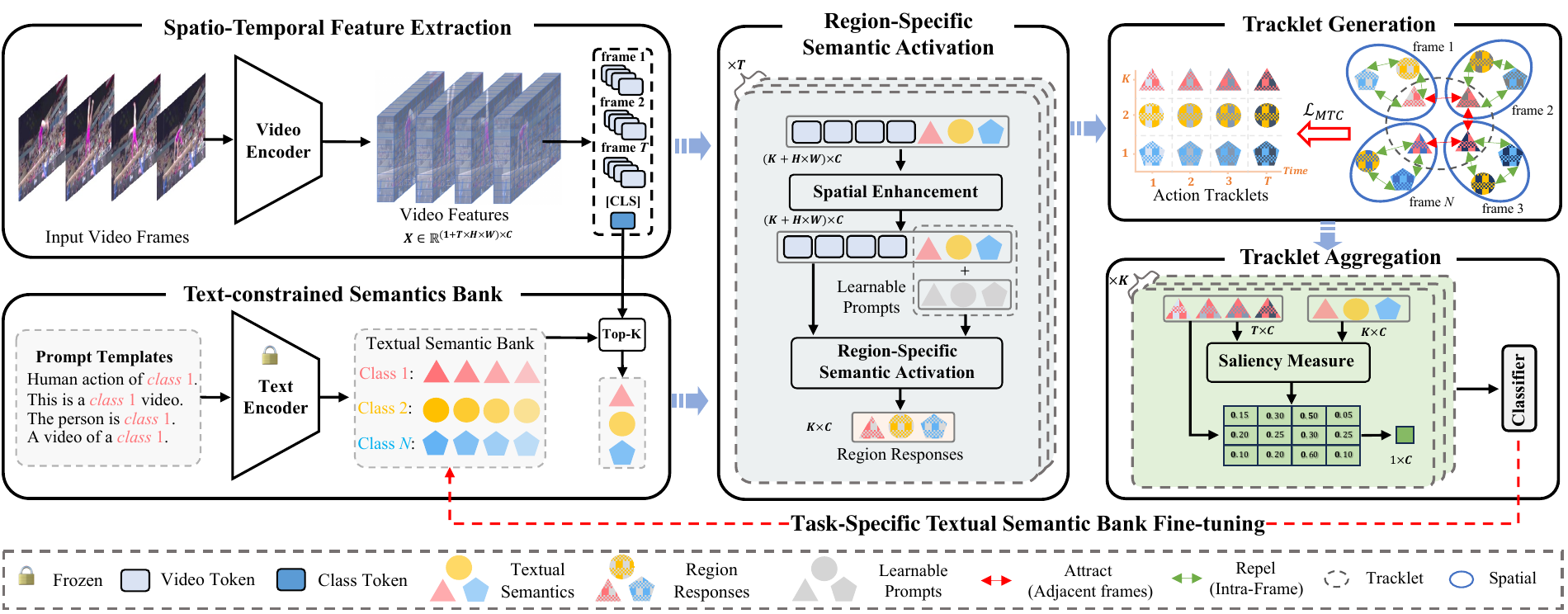}}
	\caption{\textbf{The overall framework of our proposed ART.} The backbone extracts the feature from an input video, the Spatial Semantic Extraction (SSE) component enhances the feature with the spatial context, the Region-Specific Semantics Activation (RSSA) component captures region-specific semantic responses from  enhanced region-wise representations, and the Tracklet Generation (TG) component forms a group of action tracklets, \textit{i.e.}, a group of responses to the same position queries from all video frames along the temporal dimension. 
	Finally, tracklet based representations are integrated into a global representation through Tracklet Aggergation (TA), obtaining the video's final recognition result. Furthermore, we transform action label descriptions into action phrases aligned with VLMs’ textual lexicon, forming a text-constrained semantics bank.}
	\label{fig:net}
\end{figure*}

\subsection{Contrastive Learning.}
Contrastive learning \cite{DBLP:conf/iccv/DoerschGE15} \cite{DBLP:conf/icra/SermanetLCHJSLB18} \cite{DBLP:conf/cvpr/He0WXG20} \cite{DBLP:conf/nips/GrillSATRBDPGAP20} \cite{DBLP:conf/nips/KangW0ZH20}. \cite{DBLP:conf/icml/ChenK0H20} \cite{DBLP:conf/cvpr/ChenH21} \cite{Xie_2021_CVPR} \cite{Pan_2021_CVPR} has demonstrated its great potential to learn discriminative features in a self-supervised manner. The key idea of contrastive learning lies in maximizing the similarity of representations among positive samples while exploiting discriminative patterns between positive and negative samples.
Contrastive learning can be effectively implemented by appropriately defining positives and negatives in relation to the video recognition tasks, which include augmented transformations \cite{DBLP:journals/tcsv/TaoWY22}, multi-view perspectives \cite{DBLP:conf/cvpr/WangBTT22}, and temporal coherence \cite{DBLP:journals/tip/LiuWLLL22}.
Our work departs from image-level features by learning region-based discriminative motion representations via contrastive learning, and further leveraging them to track action-specific details that enhance fine-grained action recognition.


\section{Method}

The overview of our proposed ART is shown in Figure \ref{fig:net}, consisting of five main components: (1) Spatio-Temporal Feature Extraction, (2) Region-Specific Semantics Activation, (3) Tracklet Generation, (4) Tracklet Aggregation, and (5) Text-constrained Semantics Bank. 
Firstly, we extract the spatio-temporal feature with a video encoder. Second, we introduce a set of  
text-constrained semantics  serving as region queries to  focus on regions where a specific
atomic activity occurs. Third, the identified regions of all frames are organized into action tracklets, which characterize the region-based action dynamics by linking related responses across different video frames, even without the need for supervision signals such as object proposals. Meanwhile, we design a multi-level tracklet contrastive constraint to optimize generated action tracklets. Fourth, all action tracklets are aggregated to learn the dynamic representation of distinctive local details, which are then used for the final action prediction. Lastly, in training, we refine text-constrained semantics bank  to ensure the semantic representations encoded by VLMs are not only preserved but also optimized for specific task preferences.

\subsection{Spatio-Temporal Feature Extraction}
In the feature extraction stage, we sample $T$ frames of a given video as input $ \textbf{\emph{V}}  \in \mathbb{R}^{T\times H_0\times W_0 \times 3}$, where $H_0\times W_0$ is the spatial resolution of input frames.
Following the original  UniFormer \cite{DBLP:journals/pami/LiWZGSLLQ23}, we initially apply 3D convolution to project the input video into spatio-temporal tokens. We then employ  transformer layers  for feature extraction, resulting in $1 + (T\times H\times W)$ tokens with the feature channel dimension $C$, which includes one class token and $T\times H\times W$ visual tokens (where $H\times W$ represents the spatial resolution of the features, which is 1/16 the size of the original dimensions $H_0\times W_0$). The obtained spatio-temporal feature is denoted as $\textbf{\emph{X}}  \in \mathbb{R}^{T\times H\times W \times C}$.

\subsection{Region-Specific Semantics Activation}\label{RSSA}
In order to capture the region-specific semantic responses from the visual tokens of individual frame, we propose a Region-Specific Semantics Activation (RSSA) module in this work. This module comprises a  spatial semantics enhancement stage and a region-specific semantics activation stage. 

\subsubsection{\textbf{Spatial Semantic Enhancement Stage}}
Firstly, to enhance the spatial semantics and mitigate the impact of noise in background, we introduce the Top-$K$ action-related text-constrained semantics $\textbf{\emph{S}}^{topk}  \in \mathbb{R}^{K \times C}$ from the textural semantics bank (we will provide a detailed explanation of the construction and updating of the textural semantics bank in Section III.E). Where, text-constrained semantics $\textbf{\emph{S}}^{topk}$  possess the capability to focus on the action regions. Although the textual queries are static and defined at the dataset level, their corresponding visual responses are learned dynamically, conditioned on the specific content of each video. This design allows for generalizable, semantically grounded localization without the need for video-specific text generation. As illustrated in Figure \ref{fig:net}, the spatial features of each frame ($\textbf{\emph{X}}_{t} \in \mathbb{R}^{ HW\times C}, t = 1,2,...,T$) is enhanced as follows:
\begin{equation}
\hat{\textbf{\emph{X}}}_t, \hat{\textbf{\emph{S}}}_t^{topk} = MSA(Concat(\textbf{\emph{X}}_t, \textbf{\emph{S}}^{topk})),
\end{equation}
where $Concat(\cdot)$ denotes the concatenate operation, and $MSA(\cdot)$ denotes the spatial semantic enhancement function, which is employed as multi-headed self-attention.
Then, we obtain the processed spatial feature $\hat{\textbf{\emph{X}}}_t$   that focuses more on regions relevant to the action semantics. 
Similarly, the text-constrained semantics  $\hat{\textbf{\emph{S}}}_t^{topk}$ are fine-tuned according to the specifics of the instance.

\subsubsection{\textbf{Region-Specific Semantics Activation Stage}}

As illustrated in Figure \ref{fig:net}, to obtain action-related region responses that are only activated for the regions within the frame,  the enhanced spatial feature of each frame interacts with the region queries. Inspired by DETR, we incorporate a new component which  introduces a set of learnable prompts $\textbf{\emph{P}}_{t} \in \mathbb{R}^{ K\times C}$  designed to integrate semantic representations the text-constrained semantics  $\hat{\textbf{\emph{S}}}^{topk}_t$ serving as region queries:
\begin{equation}
\textbf{\emph{Q}}_t = \textbf{\emph{P}}_{t} + \hat{\textbf{\emph{S}}}_t^{topk}.
\end{equation}

This integration allows the region queries $\textbf{\emph{Q}}_t$ to not only perceive action-related semantics but also capture individual variations within each video. We then employ the cross-attention mechanism to perform the
region-specific semantics activation. Specifically, given a spatial feature $\hat{\textbf{\emph{X}}}_t \in \mathbb{R}^{ WH\times C}$ and  a set of region queries $\textbf{\emph{Q}}_t \in \mathbb{R}^{ K\times C}$, the region-specific semantic responses are obtained through a multi-headed cross-attention operation:
\begin{equation}\label{MHCA}
    \begin{aligned}
& \quad \quad \quad  \quad  \quad \quad  \textbf{\emph{R}}_t = MCA(\textbf{\emph{Q}}_{t}, \hat{\textbf{\emph{X}}}_t) + \textbf{\emph{Q}}_{t}, \\
& MCA(\textbf{\emph{Q}}_{t}, \hat{\textbf{\emph{X}}}_t) = Softmax(\frac{(\textbf{\emph{Q}}_{t}W^Q)(\hat{\textbf{\emph{X}}}_{t}W^K)^T}{\sqrt{d}})(\hat{\textbf{\emph{X}}}_{t}W^V), \\
    \end{aligned}
\end{equation}
where $\textbf{\emph{R}}_t = \{\textbf{\emph{r}}_{t,1}, \textbf{\emph{r}}_{t,2}, ..., \textbf{\emph{r}}_{t,K}\} \in \mathbb{R}^{ K\times C}$, $MCA(\cdot)$ denotes the  multi-headed cross-attention function, the $Softmax(\cdot)$ function
is performed along $WH$ dimension, $W^Q$, $W^K$ and $W^V$ are  linear projection layers to project $\textbf{\emph{Q}}_{t}$ and $\hat{\textbf{\emph{X}}}_t$ into the latent space.

Correspondingly, the set of region-specific semantic responses for all frames $\{\textbf{\emph{R}}_1, \textbf{\emph{R}}_2, ..., \textbf{\emph{R}}_T\}$ are obtained in parallel by the $T$ RSSA modules.

\subsection{Tracklet Generation}

\subsubsection{\textbf{Tracklet Generation}}
After obtaining the region-specific semantic responses from all frames, denoted as ${\mathbf{R}_1, \mathbf{R}2, ..., \mathbf{R}T}$, we construct temporal action tracklets by temporally aligning responses corresponding to the same semantic query across frames. Each $\mathbf{R}t = {\mathbf{r}{t,1}, \mathbf{r}{t,2}, ..., \mathbf{r}{t,K}}$ contains $K$ region-specific embeddings extracted at frame $t$ via cross-attention with the set of query vectors $\mathbf{Q}_t$.
We assume that each query index $k$ corresponds to a consistent semantic concept across all frames (e.g., “arm extension” or “leg twist”). Thus, we naturally define the $k$-th action tracklet $\mathbf{Tr}k$ by temporally concatenating the $k$-th region response from every frame:
\begin{equation}
\textbf{\emph{Tr}}_k = \{\textbf{\emph{r}}_{1,k}, \textbf{\emph{r}}_{2,k}, ..., \textbf{\emph{r}}_{T,k}\},
\end{equation}

Each tracklet $\mathbf{Tr}_k$ can be interpreted as the temporal evolution of a semantically meaningful local region, governed by a fixed text-constrained query. This design provides two major advantages: (1) it avoids the need for external supervision or heuristic region association (e.g., optical flow), and (2) it enables structured temporal modeling at the semantic region level. The semantic identity of each tracklet is preserved across time by consistent indexing, while the region responses $\mathbf{r}_{t,k}$ are adaptively updated by content-specific video features during training. The resulting tracklets serve as inputs to the subsequent Multi-Level Tracklet Contrastive Loss (MTC-Loss), which refines their discriminability and temporal alignment via spatial, temporal, and tracklet-level constraints.

\subsubsection{\textbf{Multi-level Tracklet Contrastive Loss}}
Our framework obtains a fixed-size sequence of $K$ region responses for each frame through the region-specific semantics activation module. Once the tracklets are obtained, the main challenge is to constrain these tracklets to capture accurate action details without the supervision of region-level annotations. Thus, to capture diverse semantic responses in each frame and model high correlations of similar response regions between adjacent frames, we introduce a Multi-level Tracklet Contrastive Loss (MTC-Loss) among region-aware semantic responses at spatial, temporal and tracklet levels, which effectively distinguishes individual region responses of each frame (spatial level), establish correlative region responses between adjacent frames (temporal level) and generate diverse tracklets (tracklet level) in a self-supervised manner.

\noindent
\textbf{Spatial-Level Tracklet Contrastive Loss.}
Active region responses extracted from a frame should locate the discriminant differences so as to capture rich and discriminative fine-grained features. We repel the region responses in the same frame with spatial-level tracklet contrastive loss:
\begin{equation}\label{spatio}
\mathcal{L}_{spatial} =  \frac{1}{TK(K-1)}\sum_{t=1}^T\sum_{i,j=1;i\neq j}^K\frac{<\textbf{\emph{r}}_{t,i}, \textbf{\emph{r}}_{t,j}>}{||\textbf{\emph{r}}_{t,i}||_2 ||\textbf{\emph{r}}_{t,j}||_2},
\end{equation}
where $<\cdot>$ is computed based on the cosine similarity.

\noindent
\textbf{Temporal-Level Tracklet Contrastive Loss.}
Active region responses extracted from each frame of a video is not independent but correlated to an ongoing action instance collectively. We attract the responses from different frames individually, located in the same order of each frame. So the spatial-level tracklet contrastive loss is defined as:
\begin{equation}\label{temporal}
\mathcal{L}_{temporal} =  \frac{1}{T-1}\sum_{i}^K\sum_{t=1}^T\frac{<\textbf{\emph{r}}_{t,i}, \textbf{\emph{r}}_{t+1,i}>}{||\textbf{\emph{r}}_{t,i}||_2 ||\textbf{\emph{r}}_{t+1,i}||_2} - \lambda,
\end{equation}
where $\lambda$ is a hyper-parameter denoting the correlated degree of the responses from adjacent frames.

\noindent
\textbf{Tracklet-Level Tracklet Contrastive Loss.}
Furthermore, the tracklet representations should be different from each other so as to track the diverse details in a video. Our tracklet-level tracklet contrastive loss is defined as:
\begin{equation}\label{tracklet}
\mathcal{L}_{tracklet} =  \frac{1}{K(K-1)}\sum_{i,j=1;i\neq j}^K\frac{<\textbf{\emph{Tr}}_{i}, \textbf{\emph{Tr}}_{j}>}{||\textbf{\emph{Tr}}_{i}||_2, ||\textbf{\emph{Tr}}_{j}||_2}.
\end{equation}

Our multi-level tracklet contrastive loss is formed as:
\begin{equation}\label{mtc}
\mathcal{L}_{MTC} =  \mathcal{L}_{spatial} + \mathcal{L}_{temporal} + \mathcal{L}_{tracklet}.
\end{equation}

\subsection{Tracklet Aggregation and Prediction}
\subsubsection{\textbf{Tracklet Aggregation}}
\label{TA}
Mean pooling is a commonly employed method for aggregating semantic responses of each tracklet to generate the final tracklet representation.
As illustrated in Figure \ref{fig:net}, instead of treating each semantic response equally as in mean pooling, we propose a weighted aggergation that utilizes the action-related text-constrained semantics $\textbf{\emph{S}}^{topk}$
to capture the temporal saliency for tracklet aggregation.
For a set of responses $\textbf{\emph{Tr}}_k = \{\textbf{\emph{r}}_{1,k}, \textbf{\emph{r}}_{2,k}, ..., \textbf{\emph{r}}_{T,k}\}$ of $k$-th tracklet 
and a set of action-related text-constrained semantics $\textbf{\emph{S}}^{topk} = \{\textbf{\emph{s}}_{1}, \textbf{\emph{s}}_{2}, ..., \textbf{\emph{s}}_{K}\}$,
We compute the similarity between each response and each semantic embedding to assess the fine-grained relevancy. 
Next, we apply a softmax operation to normalize the similarities for each tracklet, and subsequently aggregate the similarities between a given tracklet and various semantics to derive a tracklet-level saliency.
\begin{equation}\label{agg}
    \begin{aligned}
		\mathcal{S}_{t}^{temp} = \frac{1}{K}  \sum_{i=1}^K \frac{\exp \left( <\textbf{\emph{s}}_{i},\textbf{\emph{r}}_{t,k}>\right)}{\sum_{t=1}^{T} \exp \left( <\textbf{\emph{s}}_{i},\textbf{\emph{r}}_{t,k}>\right)},\\
		\textbf{\emph{Tr}}_{k}^{agg} =  \frac{1}{T}  \sum_{t=1}^T \textbf{\emph{r}}_{t,k}\mathcal{S}_{t}^{temp}, \quad \quad \quad \quad
    \end{aligned}
\end{equation}
$\textbf{\emph{Tr}}_{k}^{agg} \in \mathbb{R}^{1\times C}$ is the final aggregated tracklet representation.

\subsubsection{\textbf{Prediction}}
A supervised task is introduced to train the backbone network for extracting task-relevant spatio-temporal feature $\textbf{\emph{X}}$ and modeling global action representation $\textbf{\emph{x}}_{cls}$ (i.e., class token):
\begin{equation}\label{Ltask}
\mathcal{L}_{v} = - \frac{1}{B} \sum_{b=1}^B ylog \ p(\textbf{\emph{x}}_{cls}),
\end{equation}
where $y$ is the action ground truth, $p(\cdot)$ is the predicted probability of a whole input video and $B$ is the number of training samples.

Finally, these tracklet representations are integrated with the global representation of the whole video for producing final recognition results:
\begin{equation}\label{Lcon}
\mathcal{L}_{con} = - \frac{1}{B} \sum_{b=1}^B y log \ p(Concat(\emph{\textbf{x}}_{cls}, \textbf{\emph{Tr}}_{1}^{agg},...,\textbf{\emph{Tr}}_{K}^{agg})).
\end{equation}

\subsection{Text-constrained Semantics Bank Generation and Fine-tuning}
\label{sec_EMA}

\begin{figure}[!h]
	\centering
	\centerline{\includegraphics[width=1\linewidth]{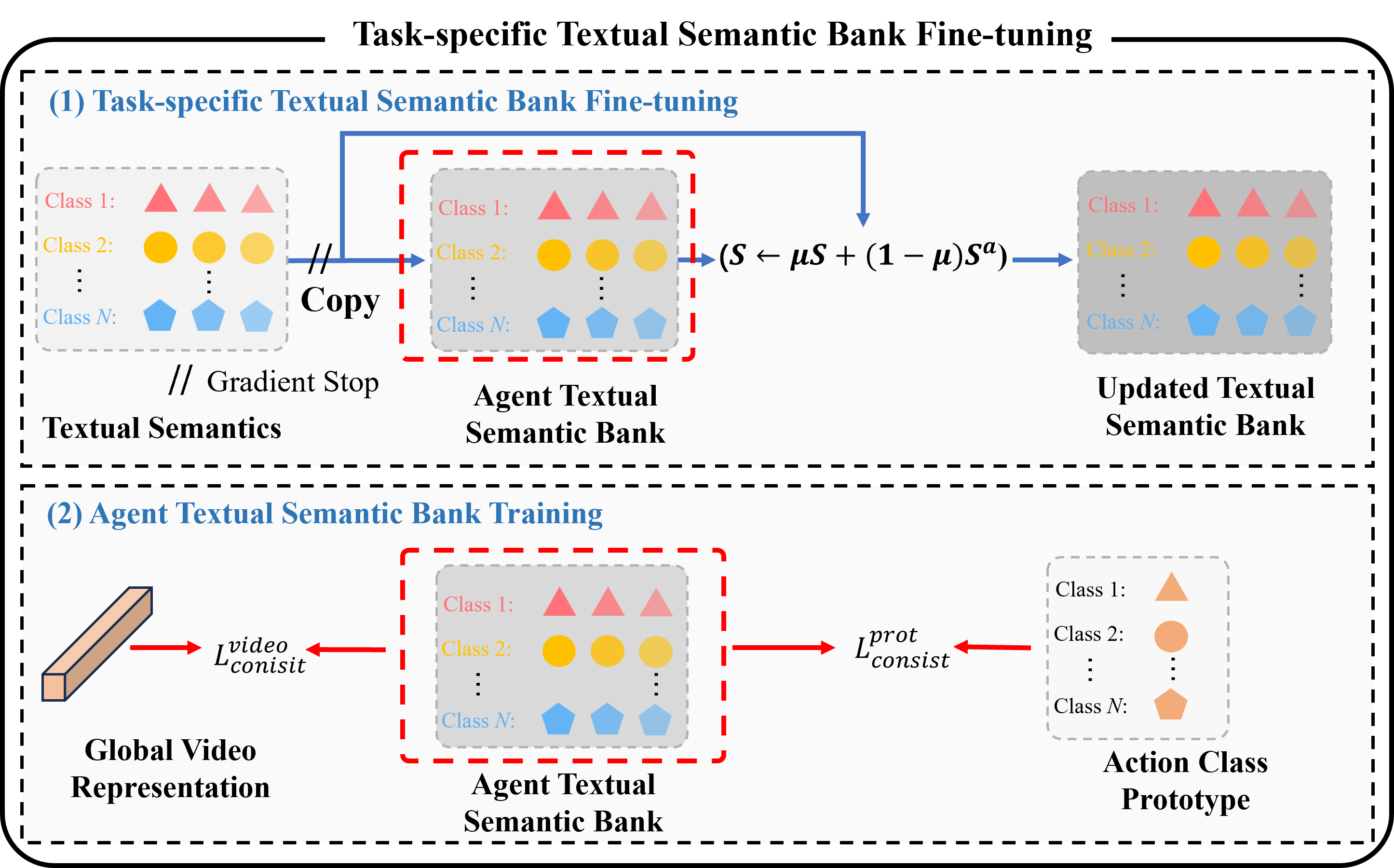}}
	\caption{\textbf{Illustration of process of task-specific textual semantic bank fine-tuning.} 
	We implement a task-specific fine-tuning mechanism for updating $\textbf{\emph{S}}$, ensuring that the textual semantics retain the semantic representations encoded by VLMs while fine-tuning to align with task-specific preferences. 
	The agent textual semantic bank $\textbf{\emph{S}}^a$ is optimized by a video consistency loss and a prototype consistency loss to narrowed the distance between categories across video and text modalities.}
	\label{fig:ema}
\end{figure}

\subsubsection{\textbf{Text-constrained Semantic Bank Generation}}
We first convert category descriptions of action labels into action phrases that approximate the pre-trained textual lexicon in Vision Language Models (VLMs). 
For example, using phrases like \textit{"This is a video about \{action label\}"} or \textit{"The person is doing \{action label\}"}, we generate $N_{prom}$ prompt templates for each action category.
Next, we utilize the pre-trained text encoder branch of CLIP~\cite{DBLP:conf/icml/RadfordKHRGASAM21} to extract semantic features from these prompt templates, initializing the textual semantic bank $\textbf{\emph{S}}^0 \in \mathbb{R}^{N_{prom} \times N_{class} \times C_t}$, where $N_{class}$ is the number of classes to classify, $C_t$ is the dimension of textual semantics.
$\textbf{\emph{S}}^0$ is then updated through an Exponential Moving Average (EMA)-like process, fine-tuning the text-constrained semantic bank to produce a highly distinct semantic bank $\textbf{\emph{S}} \in \mathbb{R}^{N_{prom} \times N_{class} \times C_t}$.

\subsubsection{\textbf{Text-constrained Semantic Bank Fine-tuning}}
We implement a task-specific fine-tuning mechanism for updating $\textbf{\emph{S}}^0$, ensuring that the textual semantics retain the semantic representations encoded by VLMs while fine-tuning to align with task-specific preferences. 
As shown in Figure ~\ref{fig:ema}, after extracting the initial textual semantic bank $\textbf{\emph{S}}^0$, we duplicate it as an agent textual semantic bank $\textbf{\emph{S}}^a$ and then truncate the gradients of $\textbf{\emph{S}}$. 
The task-specific textual semantic fine-tuning is formulated as:
\begin{equation}
\begin{aligned}
\textbf{\emph{S}}^{a^{\prime}} &\leftarrow \operatorname{optimizer}
\left(\textbf{\emph{S}}^{a}, \nabla_{\textbf{\emph{S}}^{a}} \mathcal{L}_{\mathrm{sema}}, \eta\right), \\
\textbf{\emph{S}} &\leftarrow \mu \textbf{\emph{S}} + (1-\mu)\textbf{\emph{S}}^{a^{\prime}},
\end{aligned}
\end{equation}
where $\eta$ is the learning rate, and $\mu$ is the momentum update rate, $\mathcal{L}_{sema}$ (in E.q (\ref{sema})) is the optimization function for training $\textbf{\emph{S}}_t^a$.

\textbf{Agent Textual Semantic Bank Training.}
\label{sec4}
\textbf{(1) Video Consistency Loss.} 
In the textual semantic bank, each class is represented by a total of $N_{prom}$ semantics according to $N_{prom}$ prompt templates. Therefore, video consistency loss $\mathcal{L}_{consist}^{video}$ is used to ensure the cosine consistency between the video representation $\textbf{\emph{x}}_{cls}$ and the textual semantics corresponding to the respective category. Specifically, alignment is implemented through winner-take-all classification:

\begin{align}\label{Mscore}
	\mathcal{S}_{cls} = \{(i,\mathcal{S}_{cls}^i ) | \mathcal{S}_{cls}^i  = \max_{i = 1}^{N_{prom}} \langle \  \textbf{\emph{x}}_{cls} , \ \textbf{\emph{S}}^a_{i,j} \ \rangle \}_{j=1}^{N_{class}},\\
\mathcal{L}_{consist}^{video} = -  y \cdot \log \frac{\exp \left(\mathcal{S}_{cls}\right)}{\sum_{i=1}^{N_{class}} \exp \left(\mathcal{S}_{cls}^i\right)},\quad
\end{align}
where $\mathcal{S}_{cls} \in \mathbb{R}^{1 \times {N_{class}}}$ is a classification score.

\textbf{(2) Prototype Consistency Loss.}  Prototype consistency loss $\mathcal{L}_{consist}^{prot}$  is devised to ensure the cosine consistency between the action prototypes $\mathcal{W}$ and the textual semantics corresponding to the respective category, which can be understood as the center cluster of each class, i.e., class prototype. The class prototypes  $\mathcal{W} \in \mathbb{R}^{N_{class} \times D}$ is decoupled from the prediction head. $\mathcal{L}_{consist}^{prot}$ is defined as:
\begin{align}
	\mathcal{S}_{prot} = \frac{1}{N_{prom}} \sum_{i=1}^{N_{prom}} \langle \ {MLP}(\mathcal{W}) , \ \textbf{\emph{S}}^a_i \ \rangle,  \\
    \mathcal{L}_{consist}^{prot} = - \sum_{i=1}^{N_{class}}  y_i \cdot \log \frac{\exp \left(\mathcal{S}_{prot}^i\right)}{\sum_{j=1}^{N_{class}} \exp \left(\mathcal{S}_{prot}^{i,j}\right)}, 
\end{align}
where $\mathcal{S}_{prot} \in \mathbb{R}^{N_{class} \times N_{class}}$ is a similarity matrix. Therefore, the ultimate objective function for training the agent textual semantic bank is denoted as:
\begin{equation}\label{sema}
    \mathcal{L}_{sema} = \mathcal{L}_{consist}^{video} + \mathcal{L}_{consist}^{prot}.
\end{equation}

Our ART is trained using the total loss function:
\begin{equation}\label{L}
\mathcal{L} = \mathcal{L}_{v} + \mathcal{L}_{con} + \gamma_1\mathcal{L}_{MTC} + \gamma_2\mathcal{L}_{sema}.
\end{equation}
where $\mathcal{L}_{v}$ and $\mathcal{L}_{con}$ are supervised learning losses for the global representation based prediction and the final prediction, respectively, $\mathcal{L}_{MTC}$ denotes multi-level tracklet contrastive loss. $\mathcal{L}_{sema}$ denotes text-constrained semantic bank fine-tuning loss.
We empirically set the hyperparameters $\gamma_{1}=\gamma_{2}=5$ in our experiments so that none of the loss terms dominates the training.
During the inference, action predictions are generated by the final prediction (in E.q (\ref{Lcon})).

\begin{table*}[!t]
	\begin{center}
	    \vspace{-4pt}
		\caption{Comparison of recognition performance on FineGym99, FineGym288 and Diving48. Averaged per-class accuracy (Mean\%), Top-1 accuracy (\%) and FLOPs are shown. \textcolor{red}{RED}/\textcolor{blue}{BLUE} indicate SOTA/the second best.}
		 \fontsize{8.0}{13}\selectfont
		\label{table:comp1}
		\scalebox{0.98}{
			\begin{tabular}{cccccc|cc|cc|cc}
				\toprule
				\multirow{2}{*}{\textbf{Method}}   & \multirow{2}{*}{\textbf{Backbone}} & \multirow{2}{*}{\textbf{Pretrained}}&
				\multirow{2}{*}{\textbf{Input modality}}
				&\multirow{2}{*}{\textbf{\# Frames}} &\multirow{2}{*}{\textbf{GFLOPs}}
				& \multicolumn{2}{c|}{\textbf{FineGym99}}
				&\multicolumn{2}{c|}{\textbf{FineGym288}}
                & \multicolumn{2}{c}{\textbf{Diving48}}\\
                \cline{7-12}&&&&&& \textbf{Mean}&\textbf{Top-1}&\textbf{Mean}&\textbf{Top-1}&\textbf{Mean}&\textbf{Top-1}\\
				\bottomrule
                S3D \cite{DBLP:conf/eccv/XieSHTM18} &ResNet-50 &K400&Video&8&66&72.9&81.5&42.4&75.8&36.3&50.6 \\
                TSM \cite{DBLP:conf/iccv/LinGH19} &ResNet-50 &ImageNet&Video&16&65&84.3&86.9&40.9&81.1&63.0&72.2\\
                TEA \cite{DBLP:conf/cvpr/LiJSZKW20} &ResNet-50&ImageNet&Video&16 &70&85.0&88.6&45.3&81.4&75.2&83.5\\
                TDN \cite{DBLP:conf/cvpr/0002TJW21} &ResNet-50&ImageNet&Video&16 &72&85.7&89.5&45.6&82.1
                &77.2&84.0\\ 
                 DSTS\cite{DBLP:conf/eccv/LiFKRWWL22} &ResNet-50&ImageNet&Video&16 &-&-&-&-&-
                &71.5&83.0\\
                MDCN \cite{DBLP:conf/mm/SunYYWLW22} &ResNet-50&ImageNet&Video&16 &70&\textbf{86.6}&\textbf{89.9}&\textbf{48.0}&\textbf{83.4}
                &\textbf{77.7}&\textbf{85.3}\\\hline
                \rowcolor{lightgray!40} ART &ResNet50 &ImageNet&Video&16&72 & \textcolor{red}{\textbf{87.8}}&\textcolor{red}{\textbf{91.2}}&\textcolor{red}{\textbf{48.8}}&\textcolor{red}{\textbf{84.4}}&\textcolor{red}{\textbf{78.0}}&\textcolor{red}{\textbf{85.4}}\\ \bottomrule \bottomrule
                ViViT-L \cite{DBLP:journals/corr/abs-2103-15691} &ViT-L/14&ImageNet&Video&16 &903&86.4&89.5&47.2&83.4
                &75.8&82.3\\
                TimeSformer \cite{DBLP:conf/icml/BertasiusWT21} &ViT-L/14&K00&Video&16 &1703&85.9&89.0&46.5&82.5
                &-&81.0\\
                MTV \cite{DBLP:conf/cvpr/YanXALZ0S22} &ViT-L/14&ImageNet&Video&16 &1415&86.8&89.4&47.5&83.6
                &-&83.5\\
                BEVT-V \cite{DBLP:conf/cvpr/WangCWCDLJZY22} &Swin-B&K400&Video&16 &321&-&-&-&-
                &-&83.7\\
				ActionCLIP \cite{DBLP:journals/tnn/WangXMLJ25} &ViT-B/16&K400&Video + Text&16 &1407&86.1&88.7&46.9&83.3
                &-&-\\
				X-CLIP \cite{DBLP:conf/eccv/NiPCZMFXL22} &ViT-L/14&K400&Video + Text&16 &3080&86.6&89.1&47.7&84.5
                &-&-\\
				Wu \textit{et. al.} \cite{DBLP:conf/aaai/WuSO23} &ViT-L/14&K400&Video + Text&16 &1657&86.8&90.5&49.0&85.4
                &-&-\\
                TQN \cite{DBLP:conf/cvpr/Zhang0Z21} &ViT-L/14&K400&Video&\textbf{48} &-&\textcolor{blue}{\textbf{90.6}}&\textcolor{blue}{\textbf{93.8}}&\textcolor{red}{\textbf{61.9}}&\textcolor{blue}{\textbf{89.6}}&74.5&81.8\\
                UniFormerV2 \cite{DBLP:juniformerv2iccv23} &ViT-L/14&K400&Video&16 &1332&86.4&92.9&52.4&87.2&\textcolor{blue}{\textbf{78.4}}&\textcolor{blue}{\textbf{85.9}}\\\hline 
                \rowcolor{lightgray!40} ART  &ViT-B/16&K400&Video + Text&16&436&89.2&93.9&54.6&89.9&78.8&87.0\\
                \rowcolor{lightgray!40} ART  &ViT-L/14&K400&Video + Text&16&1390&\textcolor{red}{\textbf{90.9}}&\textcolor{red}{\textbf{94.7}}&\textcolor{blue}{\textbf{56.1}}&\textcolor{red}{\textbf{90.2}}&\textcolor{red}{\textbf{80.2}}&\textcolor{red}{\textbf{87.9}}\\
                \bottomrule
		\end{tabular}}
	\end{center}
\end{table*}

\section{Experiments and Discussions}

\subsection{Implementation Details}
We use CNN-based TEA \cite{DBLP:conf/cvpr/LiJSZKW20} and Transformer-based UniFormerV2 \cite{DBLP:juniformerv2iccv23}  as the backbones for feature extraction, respectively. 
We evaluate the performance of ART on four widely-used action recognition datasets. \textbf{FineGym \cite{DBLP:conf/cvpr/ShaoZDL20a}} released very recently is a large-scale highquality action dataset with fine-grained annotations, which was collected from HD gymnasium videos of 288 categories. \textbf{Diving48 \cite{DBLP:conf/eccv/LiLV18}} contains 48 categories with 16k training\/2k testing of competitive diving videos. \textbf{NTURGB-D \cite{DBLP:journals/pami/LiuSPWDK20}} is a large-scale RGB-D dataset with synchronized skeleton and video data, widely used for action recognition and pose estimation, and includes NTU-60 and NTU-120 subsets. \textbf{Kinetics-400/-600 \cite{DBLP:conf/cvpr/CarreiraZ17}} is a widely used large-scale benchmark dataset for action recognition, consisting of 55 hours of recordings capturing daily activities.
The training parameters for all the datasets are: 50 training epochs, initial learning rate 0.01 (decreased by 0.1 for every 20 epochs), dropout rate 0.5. 
The sparse sampling strategy \cite{DBLP:conf/cvpr/CarreiraZ17} is used to extract $T$ frames from videos ($T=16$). 
Each input frame is cropped and resized to $224\times224$ for training and testing. Random scaling, cropping and horizontal flipping are deployed as data augmentation. 
We use standard transformer layer \cite{DBLP:conf/nips/VaswaniSPUJGKP17} with four layers  in spatial enhancement and region-specific semantics activation, respectively.
The dimension of the latent space $d$ is 256, and the number of queries of each frame $K$ (\textit{a.k.a.}, the number of tracklets) is set as 2.
The chosen hyperparameters in our approach are $\gamma_{1}=\gamma_{2}=5$ and $\lambda = 0.6$.
The ablation studies presented below will demonstrate the effectiveness of our configurations.

\subsection{Comparison with the State-of-the-Art}

As our work is focused on fine-grained action recognition, the evaluation is mainly conducted on two publicly available fine-grained action dataset, FineGym \cite{DBLP:conf/cvpr/ShaoZDL20a}, Diving48 \cite{DBLP:conf/eccv/LiLV18} and NTURGB-D \cite{DBLP:journals/pami/LiuSPWDK20}.
As shown in Table \ref{table:comp1}, we provide the overall performance comparison  between our proposed ART and other state-of-the-art methods
in terms of two metrics, per-class accuracy (Mean\%) and per-video  accuracy (Top-1\%).
Quantitative results are listed in Table \ref{table:comp1}, where models are grouped into two categories: CNN-based and Transformer-based. From these results of CNN-based models, we can see that the overall performance of ART outperforms the most competitive existing method, MDCN, on all three datasets.
From these results of Transformer-based models, our ART achieves 94.7\%, 90.2\% and 87.9 Top-1 accuracies, respectively, improving UniFormerV2 by 1.8\%, 3.0\% and 2.0\% on all three datasets. Our ART only falls behind TQN in the Mean accuracy on the FineGym288 dataset. TQN was introduced using a transformer architecture to facilitate the learning of granularity across different attributes, guided by the supervision of multi-attribute sub-labels derived from multipart text descriptions of action labels. TQN also used more input frames (\textit{i.e.}, 48 frames over the popular setting 16 frames) to achieve Mean accuracy 61.9\%. Our ART achieves competitive results with fewer input frames.
These experimental results clearly demonstrates the effectiveness of our action tracklet network for fine-grained action recognition.
We compare our ART with other video+text baselines in Table I. Notably, although ActionCLIP, X-CLIP, and Wu et al. utilize similar external textual supervision, they still fall behind our ART model in all metrics across datasets. For instance, on FineGym99, our ViT-L/14-based ART achieves 90.9\% Top-1 accuracy, significantly outperforming X-CLIP (89.6\%) and Wu \textit{et al.} (88.9\%). Similar trends are observed on Diving48 and FineGym288. These results demonstrate that the performance gain is not simply due to access to text data, but stems from our task-aligned tracklet-based architecture and the MTC loss, which better leverage textual semantics in a temporally consistent and fine-grained manner.
In addition to the accuracy metrics, we also report inference cost in terms of FLOPs. It is noticed that our ART has similar computational costs due to the adoption of the same backbones. Since we use a lightweight transformer architecture, ART's FLOPs is still at a low level.

We further assessed the performance of our ART on the NTU-60 and NTU-120 datasets, analyzing the cross-subject (XSub), cross-view (XView), and cross-setup (XSet) scenarios individually for each dataset. 
Table \ref{table:comp2} showcases the recognition performance comparison  across different video classification benchmarks. The results underscore the consistent superiority of ART over established baselines.



\begin{table}[!t]
	\begin{center}
	\caption{Comparison of recognition accuracy on the NTURGB-D with Cross-Subject (XSub), Cross-View (XView), and Cross-Setup (XSet) settings in terms of Top-1 accuracy (\%). \textcolor{red}{RED}/\textcolor{blue}{BLUE} indicate SOTA/the second best.}
			 \fontsize{8.0}{13}\selectfont
			\label{table:comp2}
			\scalebox{0.91}{
				\begin{tabular}{ccc|cc|cc}
					\toprule
					\multirow{2}{*}{\textbf{Method}}   & \multirow{2}{*}{\textbf{Pose}}&
					\multirow{2}{*}{\textbf{Video}}
					& \multicolumn{2}{c|}{\textbf{NTU60}}
					&\multicolumn{2}{c}{\textbf{NTU120}}\\
					\cline{4-7}&&& \textbf{XView}&\textbf{XSub}&\textbf{XSet}&\textbf{XSub}\\
					\bottomrule
				 \textcolor{gray}{Separable STA \cite{DBLP:conf/iccv/DasDKMGBF19}}&\textcolor{gray}{\usym{1F5F8}}&\textcolor{gray}{\usym{1F5F8}}& \textcolor{gray}{94.6}&\textcolor{gray}{92.2}& \textcolor{gray}{82.5}&\textcolor{gray}{83.8}\\
				\textcolor{gray}{MS-G3D \cite{DBLP:conf/cvpr/LiuZC0O20}}&\textcolor{gray}{\usym{1F5F8}}&\textcolor{gray}{\usym{1F5F8}}&\textcolor{gray}{96.2} &\textcolor{gray}{91.5} &\textcolor{gray}{88.4} &\textcolor{gray}{87.2}\\
				\textcolor{gray}{Hands Attention\cite{Baradel2018HumanAR}}&\textcolor{gray}{\usym{1F5F8}}&\textcolor{gray}{\usym{1F5F8}}&\textcolor{gray}{90.6} &\textcolor{gray}{84.8} &\textcolor{gray}{-} &\textcolor{gray}{-}\\
				\textcolor{gray}{S-Res-LSTM \cite{DBLP:journals/tip/SongLLG20}}&\textcolor{gray}{\usym{1F5F8}}&\textcolor{gray}{\usym{1F5F8}}&\textcolor{gray}{96.3} &\textcolor{gray}{90.0} &\textcolor{gray}{-} &\textcolor{gray}{-}\\
				\textcolor{gray}{TSMF \cite{DBLP:conf/aaai/YuLC21}}&\textcolor{gray}{\usym{1F5F8}}&\textcolor{gray}{\usym{1F5F8}}&\textcolor{gray}{97.4} &\textcolor{gray}{92.5} &\textcolor{gray}{89.1} & \textcolor{gray}{87.0}\\
				\textcolor{gray}{STAR   \cite{DBLP:conf/wacv/AhnKHK23}}&\textcolor{gray}{\usym{1F5F8}}&\textcolor{gray}{\usym{1F5F8}}&\textcolor{gray}{96.5}&\textcolor{gray}{92.9}&\textcolor{gray}{92.7}&\textcolor{gray}{90.3}\\
				\textcolor{gray}{VPN++ \cite{9613748}} &\textcolor{gray}{\usym{1F5F8}}&\textcolor{gray}{\usym{1F5F8}} &\textcolor{gray}{98.1} &\textcolor{gray}{94.9} &\textcolor{gray}{92.5} &\textcolor{gray}{90.7}\\\hline 
				I3D+NL \cite{DBLP:conf/cvpr/CarreiraZ17} &\ding{55}&\usym{1F5F8} & -&- &34.3 &43.9\\
				GliClouds \cite{DBLP:conf/cvpr/Baradel0MT18} &\ding{55}&\usym{1F5F8} &93.2 &86.6 &- &-\\
				VPN++ \cite{9613748}&\ding{55}&\usym{1F5F8} &\textcolor{blue}{\textbf{94.9}} &\textcolor{blue}{\textbf{91.9}} &89.3 &86.7\\
				UniFormerV2-L \cite{DBLP:juniformerv2iccv23} &\ding{55}&\usym{1F5F8}&93.0&91.8&\textcolor{blue}{\textbf{90.0}}&\textcolor{blue}{\textbf{87.4}}\\\hline 
				  \rowcolor{lightgray!40} ART (ViT-L/14)  &\ding{55}&\usym{1F5F8}&\textcolor{red}{\textbf{95.2}}&\textcolor{red}{\textbf{93.1}}&\textcolor{red}{\textbf{90.5}}&\textcolor{red}{\textbf{88.7}}\\
					\bottomrule
			\end{tabular}}
		\end{center}
	\end{table}

	\begin{table}[!t]
		\begin{center}
			\caption{Recognition performance on Kinetics-400 and Kinetics-600 in terms of Top-1 accuracy (\%) and Top-5 accuracy (\%). \textcolor{red}{RED}/\textcolor{blue}{BLUE} indicate SOTA/the second best.}
			 \fontsize{8.0}{13}\selectfont
			\label{table:comp22}
			\scalebox{0.72}{
				\begin{tabular}{c|ccc|cc|cc}
					\toprule
					\multirow{2}{*}{\textbf{Method}}&
					\multirow{2}{*}{\textbf{Modality}}
					 &\multirow{2}{*}{\textbf{Frames}}
					&\multirow{2}{*}{\textbf{GFLOPs}}
					& \multicolumn{2}{c|}{\textbf{Kinetics400}}
					& \multicolumn{2}{c}{\textbf{Kinetics600}}\\
					\cline{5-8} &&&& \textbf{Top-1}&\textbf{Top-5} & \textbf{Top-1}&\textbf{Top-5}\\
					\bottomrule
				   TEA \cite{DBLP:conf/cvpr/LiJSZKW20}&Video&16&70&74.0&91.3&67.9&87.4\\
				   TDN \cite{DBLP:conf/cvpr/0002TJW21}&Video&16&72& 76.8&93.0&-&-\\
				   SlowFast-R01 \cite{DBLP:conf/iccv/Feichtenhofer0M19} &Video&32 &7020& 79.8 & 93.9 &81.8& 95.1 \\
				   X3D-XL \cite{DBLP:conf/cvpr/Feichtenhofer20} &Video&32&1452 &79.1 &93.9&81.9&95.5\\
				   TimeSformer \cite{DBLP:conf/icml/BertasiusWT21} &Video&64&7140&80.7&94.7&82.2&95.6\\
				   Video-Swin-L \cite{DBLP:conf/cvpr/LiuN0W00022}  &Video&32 &16890&83.1 & 95.9 & 85.9&97.1  \\
				   MTV-L \cite{DBLP:conf/cvpr/YanXALZ0S22} &Video&32 &18050&84.3 & 96.3&85.4&96.7 \\
				   X-CLIP-L \cite{DBLP:conf/eccv/NiPCZMFXL22} &Video + Text&8 &1868 &87.1 & 97.6 &88.3 &97.7\\
				   ActionCLIP \cite{DBLP:journals/tnn/WangXMLJ25} &Video + Text& 32&16890 & 83.8 &97.1 &-&- \\
				   SSTSA \cite{DBLP:journals/tnn/AlfaslyCJLX24} &Video &32&1356 & 83.1 & - &-&- \\
				   MAR \cite{DBLP:journals/tmm/QingZHWWLGS24} &Video &16&4140 & 85.3 & 96.3 &-&- \\
				   VideoMamba \cite{DBLP:journals/corr/abs-2403-06977} &Video&16 &2424 &83.4 &95.9 &- &- \\
				   UniFormerV2-L \cite{DBLP:juniformerv2iccv23} &Video&16&1332&\textcolor{blue}{\textbf{89.1}}&\textcolor{blue}{\textbf{98.2}} &\textcolor{blue}{\textbf{89.5}}&\textcolor{blue}{\textbf{98.3}}\\\hline
				   \rowcolor{lightgray!40} ART (ViT-L/14) &Video + Text&16 & 1340 &\textcolor{red}{\textbf{90.3}} &\textcolor{red}{\textbf{98.4}}&\textcolor{red}{\textbf{89.9}}&\textcolor{red}{\textbf{98.5}} \\\bottomrule
			\end{tabular}}
		\end{center}
	\end{table}

We also evaluate the generalization potential of our model for conventional action recognition on Kinetics-400 and Kinetics-600 which are less sensitive to fine granularity and the dynamics of distinctive local details.
As shown in Table \ref{table:comp22}, our ART outperforms other existing methods by achieving 90.3\% and 89.9\% Top-1 accuracy on Kinetics-400 and Kinetics-600 datasets.
These results clearly demonstrate the effectiveness and generalization of our model for conventional action recognition.


\begin{table*}[t]
	\begin{center}
		\caption{Impact of the key components in ART on FineGym99, Diving48 and NTU60-XView in terms of Top-1 and Mean accuracies (\%).}
		 \fontsize{8.0}{13}\selectfont
		\label{table:a1}
		\scalebox{0.90}{
			\begin{tabular}{cc|cc|cc|cc}
				\toprule
				\multirow{2}{*}{}&\multirow{2}{*}{\textbf{Setting}} &  \multicolumn{2}{c|}{\textbf{FineGym99}} &  \multicolumn{2}{c|}{\textbf{Diving48}}&  \multicolumn{2}{c}{\textbf{NTU60-XView}}\\
                   \cline{3-8} &&\textbf{Top-1}&\textbf{Mean}&\textbf{Top-1}&\textbf{Mean}&\textbf{Top-1}&\textbf{Mean}\\
				\bottomrule
                (1) &Baseline &90.5&85.7&84.4&75.4 &92.7&93.1 \\
                (2) &Baseline + RSSA &90.9 &86.2&85.2& 76.5&93.1&93.5\\
                (3) &Baseline + RSSA w/o SE &90.4 &85.8& 84.0 &75.2 &92.0&92.4\\
                (4) &Baseline + RSSA  + TA &91.7&87.3 & 85.5 &77.3 &93.7&94.0\\\hline
                \rowcolor{lightgray!40}(5) &Baseline  + RSSA + TA + MTC Loss  (ART) &\textbf{92.7} (+2.2\%)&\textbf{88.5} (+2.8\%)&\textbf{86.3} (+1.9\%) &\textbf{78.2} (+2.8\%)&\textbf{94.4} (+1.7\%) &\textbf{94.6} (+1.5\%) \\\bottomrule
		\end{tabular}}
	\end{center}
\end{table*}

\begin{table*}[t]
	\begin{center}
		\caption{Impact of each component in MTC-Loss on FineGym99, Diving48 and NTU60-XView in terms of Top-1 and Mean accuracies (\%).}
		 \fontsize{8.0}{13}\selectfont
		\label{table:mtc}
		\scalebox{0.90}{
			\begin{tabular}{cccc|cc|cc|cc}
				\toprule
				\multirow{2}{*}{} & \multirow{2}{*}{\textbf{$\mathcal{L}_{spatial}$}} & \multirow{2}{*}{\textbf{$\mathcal{L}_{temporal}$}} & \multirow{2}{*}{\textbf{$\mathcal{L}_{tracklet}$}}&  \multicolumn{2}{c|}{\textbf{FineGym99}} &  \multicolumn{2}{c|}{\textbf{Diving48}} &  \multicolumn{2}{c}{\textbf{NTU60-XView}}\\ 
               \cline{5-10} &&&&\textbf{Top-1}&\textbf{Mean}&\textbf{Top-1}&\textbf{Mean}&\textbf{Top-1}&\textbf{Mean}\\ \bottomrule
                (1)& & & &91.7&87.3 & 85.5 &77.3 &93.7&94.0 \\
                (2)& \usym{1F5F8} & & &92.1 &87.7 &85.7 &77.4 &93.8 &94.3\\
                (3)& & \usym{1F5F8} & & 90.9 &86.8&85.0 &77.1 &92.1 & 93.0\\
                (4)& & & \usym{1F5F8}&91.9&87.8&85.9&77.6 &94.0 & 94.2\\ 
                (5)& \usym{1F5F8}&\usym{1F5F8} & &92.5 &88.2&86.0&77.9 & 94.2 & 94.5\\\hline
                \rowcolor{lightgray!40}(6)& \usym{1F5F8}& \usym{1F5F8} & \usym{1F5F8} &\textbf{92.7} (+1.0\%)&\textbf{88.5} (+1.2\%)&\textbf{86.3} (+0.8\%) &\textbf{78.2} (+0.9\%)&\textbf{94.4} (+0.7\%) &\textbf{94.6} (+0.6\%)\\
				\bottomrule
		\end{tabular}}
	\end{center}
\end{table*}

\begin{table*}[t]
	\begin{center}
		\caption{Impact of each component in task-specific text-constrained semantic bank fine-tuning on FineGym99, Diving48 and NTU60-XView in terms of Top-1 and Mean accuracies (\%).}
		 \fontsize{8.0}{13}\selectfont
		\label{table:ab3}
		\scalebox{0.98}{
			\begin{tabular}{ccc|cc|cc|cc}
				\toprule
				\multirow{2}{*}{} & \multirow{2}{*}{\textbf{$\mathcal{L}_{consist}^{video}$}} & \multirow{2}{*}{\textbf{$\mathcal{L}_{consist}^{prot}$}} &  \multicolumn{2}{c|}{\textbf{FineGym99}} &  \multicolumn{2}{c|}{\textbf{Diving48}} &  \multicolumn{2}{c}{\textbf{NTU60-XView}}\\ 
               \cline{4-9} &&&\textbf{Top-1}&\textbf{Mean}&\textbf{Top-1}&\textbf{Mean}&\textbf{Top-1}&\textbf{Mean}\\ \bottomrule
                (1)&  & &92.7&88.5 & 86.3 &78.2 &94.4&94.6 \\
                (2)& \usym{1F5F8} &  &92.9 &88.6 &86.5 &78.6 &94.6 &95.0\\
                (3)& & \usym{1F5F8} & 93.1 &88.8&86.9 &78.5 & 94.4&94.3\\\hline
                \rowcolor{lightgray!40}(4)& \usym{1F5F8}& \usym{1F5F8}  &\textbf{93.5} (+0.8\%)&\textbf{89.2} (+0.7\%)&\textbf{87.0} (+0.7\%) &\textbf{78.8} (+0.6\%)&\textbf{94.7} (+0.3\%) &\textbf{95.6} (+1.0\%)\\
				\bottomrule
		\end{tabular}}
	\end{center}
\end{table*}

\subsection{Ablation Study}

We conduct extensive ablation studies to evaluate the impact of various design choices, including  spatial enhancement (SE) stage, region-specific semantics activation (RSSA) module and tracklet aggregation (TA) module. 
Ablation studies are conducted on FineGym99, Diving48 and NTU60-XView, using UniFormerV2 (ViT-B/16) as the  backbone. 
As shown in  Table \ref{table:a1}, the experiments are conducted under five settings: 
(1) Baseline: The baseline starts from the backbone network  by simply feeding the backbone's output features of a given video into a classifier for action prediction; 
(2) Baseline + RSSA:  The output features of the backbone are fed into the RSSA module to obtain region-specific semantic responses, which are then arranged into tracklets.
In this case, we used a simple strategy of directly adding the elements to aggregate the tracklets. The aggregated tracklet representation is directly concatenated with the video class token $\textbf{\emph{x}}_{cls}$ for final prediction;
(3) Baseline + RSSA (w/o SE):  To verify the effectiveness of the spatial enhancement (SE) stage in the RSSA module, we conducted experiments by removing the SE.
(4) Baseline + RSSA + TA: Unlike Setting (2), we performed tracklet aggregation (TA) module to merge all the tracklets.
(5) Baseline + RSSA + TA w/ MTC Loss: For generated tracklets, we introduce MTC Loss to constrain these tracklets to capture action details accurately.

\begin{figure}[htbp]
	\centering
	\centerline{\includegraphics[width=1\linewidth]{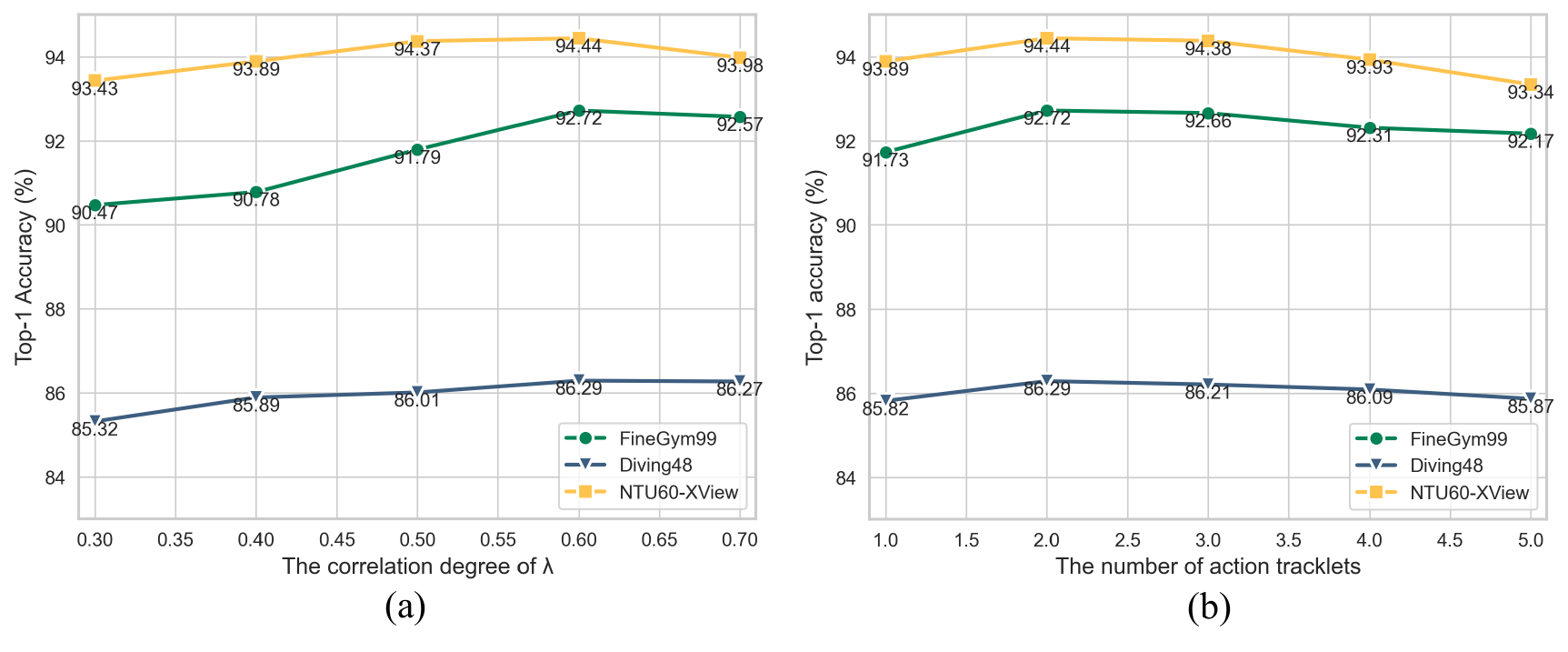}}
	\caption{Impact of (a) the correlation degree of $\lambda$ and (b) the number of action tracklets on FineGym99, Diving48 and NTU60-XView in terms of Top-1 accuracy}
	\label{fig:ab_number}
\end{figure}

\begin{figure}[htbp]
	\centering
	\centerline{\includegraphics[width=1\linewidth]{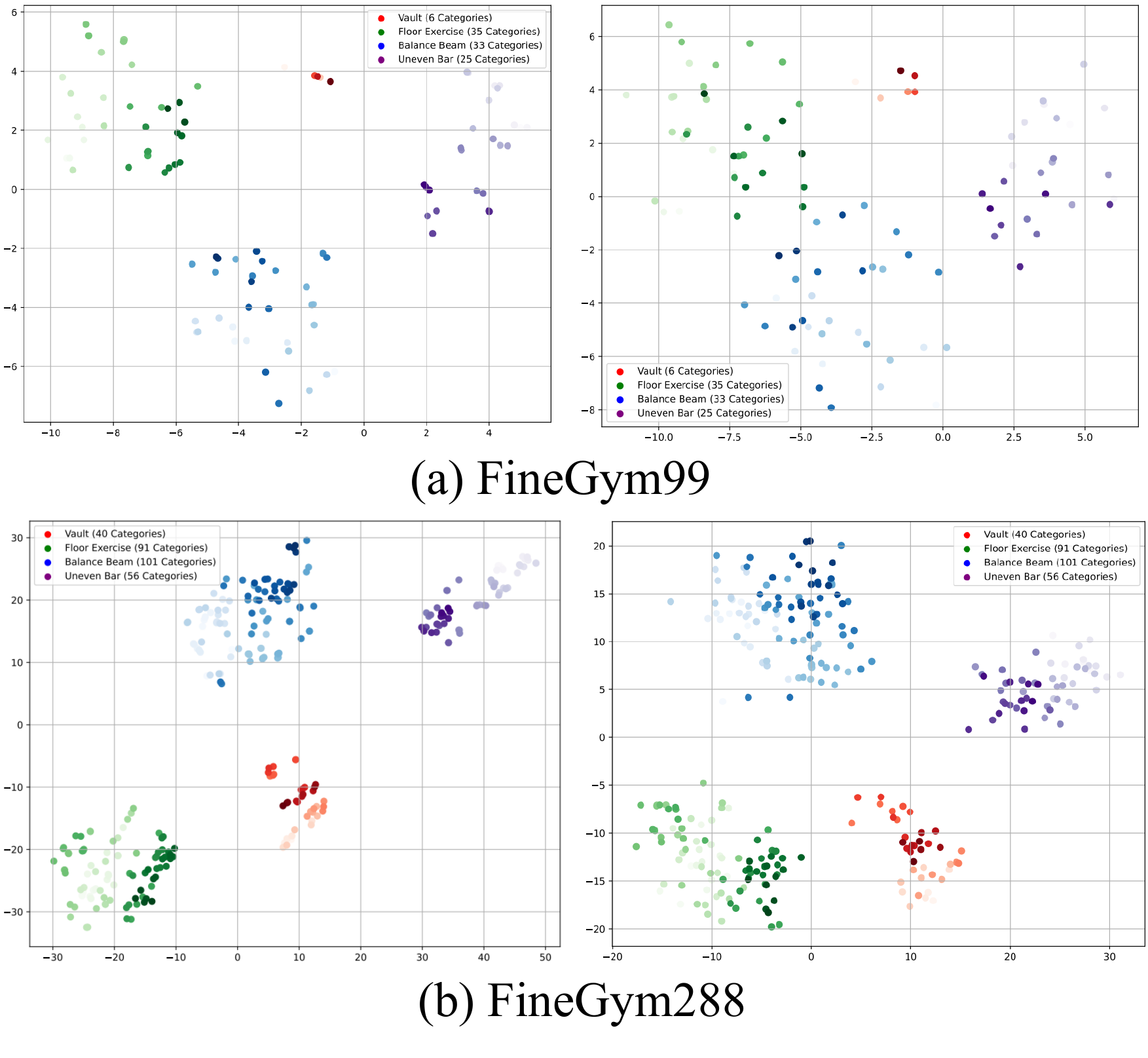}}
	\caption{The t-SNE visualization results without (\textbf{left}) and with (\textbf{Right}) task-specific fine-tuning. 
	(a) and (b) inducate the visualizations on FineGym99 and FineGym288, respectively. 
	These results show that our task-specific fine-tuning mechanism significantly enlarges the inter-class distribution differences.}
	\label{fig:vis2}
\end{figure}

\textbf{Effectiveness of SE.} From Table \ref{table:a1}, we can observe significant performance improvement of Setting (2) over the baseline Setting (3), improved by 0.5\%, 0.8\% and 0.9 Top-1 accuracies on FineGym99, Diving48 and NTU60-XView, respectively. 
This clearly shows that our spatial enhancement, \textit{i.e.}, the encoder architecture, is able to capture 
saliency semantics in spatial  dimension and is more effective for fine-grained action recognition, compared to the backbone.

\textbf{Effectiveness of Action Tracklet (RSSA and TA).}
As described in Sec. \ref{RSSA} and Sec. \ref{TA}, we obtain region-specific semantic responses through RSSA and aggregate action tracklets through TA for the final prediction. 
To demonstrate the effectiveness of the action tracklet, we implement our model under Setting (2) and Setting (4) in Table \ref{table:a1}, where the only difference is the usage of our proposed TA module. 
It can be observed that Setting (4) gains a significant performance improvement compared to the baseline. This fully verifies the importance of considering region-based action dynamics for action recognition.
Moreover, Setting (4) also achieves a performance improvement over Setting (2).This demonstrates that simply generating action tracklets  may not capture the discriminative ones. 
After introducing the TA module, better performance is achieved.

We also evaluate the impact of the number of action tracklets under Setting (4) in Figure \ref{fig:ab_number} (a). 
It is also equivalent to the number of the discriminative regions of interest in a frame. 
As the objects of interest and background in fine-grained action videos are generally the same or very similar, discriminative action-specific differences only occur in several regions. 
We investigate the effect of varying the number of action tracklets $K$, and find that the best performance is achieved when when $K=2$.

\textbf{Effectiveness of MTC Loss. }
We investigate the performance of our ART with different components of MTC Loss. 
As shown in Table \ref{table:mtc}, the final case with all three loss terms consistently outperforms the other cases. 
Starting form the baseline (the first case), adding $\mathcal{L}_{spatial}$ or $\mathcal{L}_{tracklet}$ alone can improve the performance, respectively. 
When they are integrated together (the fifth case),  the performance increases further. 
We should note that using $\mathcal{L}_{temporal}$ alone may decrease the performance (even lower than that of the baseline), however introducing it together with both spatial and tracklet losses (the last case) can boost the performance further. 
It demonstrates that  only when the meaningful responses are discovered at intra-frame  level, $\mathcal{L}_{temporal}$ can work well to link these responses together as the tracklets. 
To conclude, MTC loss can help capture action details in each frame accurately and track the active regions over time, thereby further enhancing discriminative representation for fine-grained action recognition.

\begin{figure}[htbp]
	\centering
	\centerline{\includegraphics[width=0.97\linewidth]{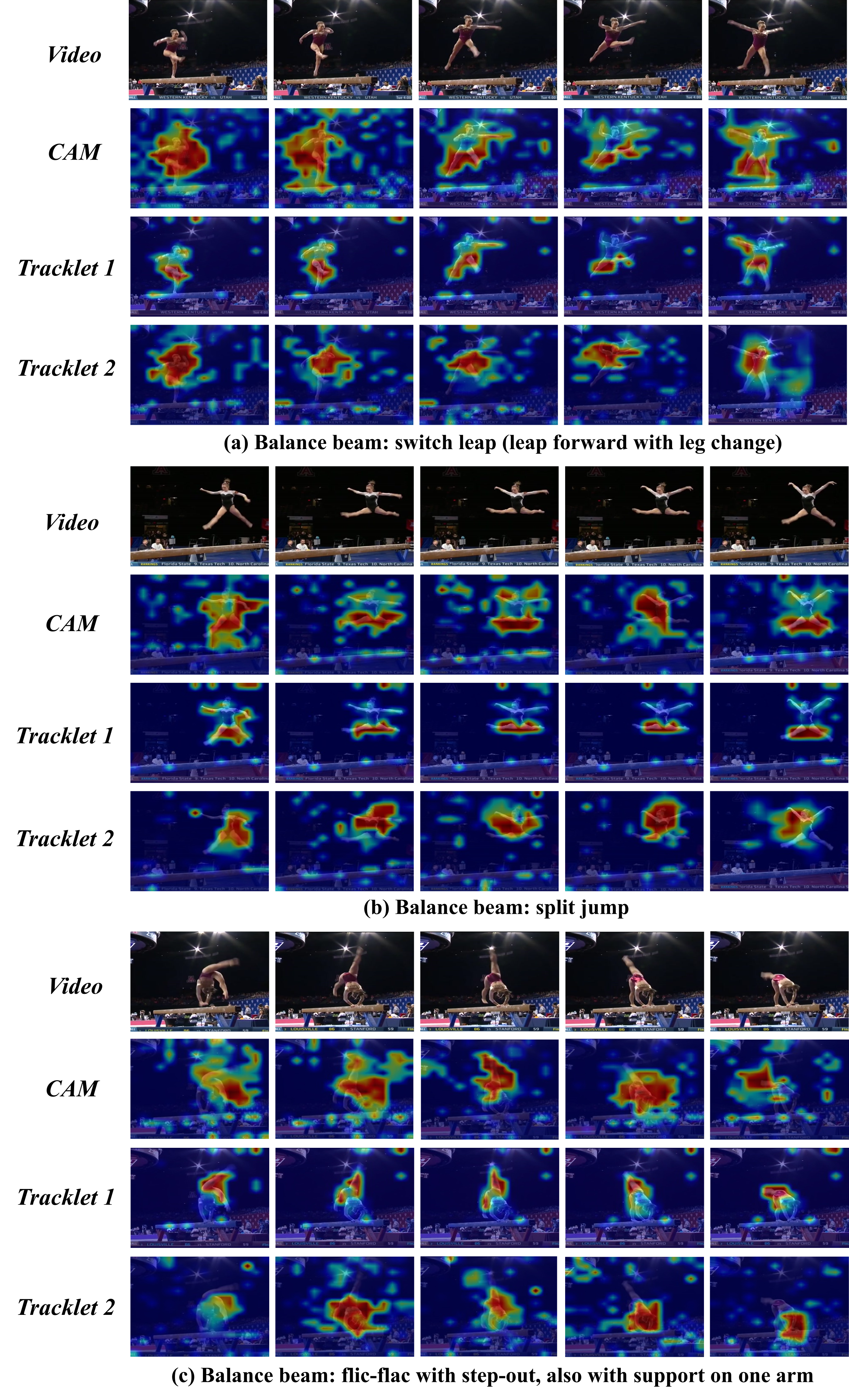}}
	\caption{\textbf{Visualization of activation maps comparing baseline CAM and  region-specific semantic responses on three action samples of FineGym288}. 
	(a) Switch leap: ART attends to the leading leg and hip, effectively capturing the transition phase that differentiates leap types.
(b) Split jump: Attention concentrates on thigh and foot extensions, modeling the dynamics of leg separation.
(c) Flic-flac with step-out (one-arm support): ART exhibits partial misalignment, with incomplete tracking of the supporting arm under high-speed motion.}
	\label{fig:vis}
\end{figure}

In Figure \ref{fig:ab_number} (b), we evaluate the impact of the correlation degree $\lambda$ of $\mathcal{L}_{temporal}$.
$\lambda$ denotes the correlation degree between neighbouring frames, it does affect the quality of action tracklets.
Due to limb deformation and view change for adjacent frames, $\lambda$ is set to 0.6  to obtain the best results supported by experimental validation.

\begin{table}[t]
	\begin{center}
	\caption{Comparison of computational complexity, model parameter count and recognition accuracy.}
		 \fontsize{8.0}{13}\selectfont
	  	\label{table:GFLOPs}
		\scalebox{0.585}{
			\begin{tabular}{c|c|c|c|c|c}
				\toprule
				\textbf{Method} & \textbf{GFLOPs} & \textbf{Param (M)} & \textbf{Infer. Time (ms)}& \textbf{Epoch Ave. Time (min)} & \textbf{Mean (\%)} \\
				\bottomrule
                 UniformerV2 \cite{DBLP:juniformerv2iccv23}  &1332.43 & 345.05 & 110.40& 90.5 &85.7\\           
                ART (Ours) &1429.32 ($\uparrow$7.27\%)& 365.01 ($\uparrow$5.78\%) & 36.2 ($\uparrow$1.11\%) &121.82 ($\uparrow$10.3\%) &89.2 ($\uparrow$3.5\%)\\ \bottomrule
		\end{tabular}}
	\end{center}
\end{table}

\textbf{Effects of the Task-specific Textual Semantic Fine-tuning}. In our investigation, we focused on the fine-tuning mechanism of the textual semantic bank $\mathcal{S}$, specifically employing an exponential moving average (EMA) approach supervised by cosine similarity loss to advance textual semantics for cross-modality alignment.
As a result, substantial performance improvements were observed on the FineGym99, Diving48, and NTU60-View datasets compared to baseline. 
We further validate the effectiveness of our proposed video consistency loss $\mathcal{L}_{consist}^{video}$  and prototype consistency loss $\mathcal{L}_{consist}^{prot}$.
Under the supervision of $\mathcal{L}_{consist}^{video}$  and  $\mathcal{L}_{consist}^{prot}$, the textual semantic bank updated  strategy not only established stronger consistency between different textual semantics of the same category but also narrowed the distance between categories across video and text modalities. 
Comparative results indicate that,  the implementation of consistency losses led to a more notable enhancement in performance. 

To illustrate that the task-specific Textual Semantic Fine-tuning Mechanism effectively update the initial textual semantic, 
we first visualize the semantic embeddings of textual semantic bank using t-SNE without (left) and
with (Right) task-specific fine-tuning. 
The embeddings are more dispersed compared to those of the baseline, which indicates that our method has the ability to extract more discriminative category semantics.
This significantly enlarges the inter-class distribution differences, thereby improving performance of fine-grained action recognition. 

\begin{figure}[htbp]
	\centering
	\centerline{\includegraphics[width=0.97\linewidth]{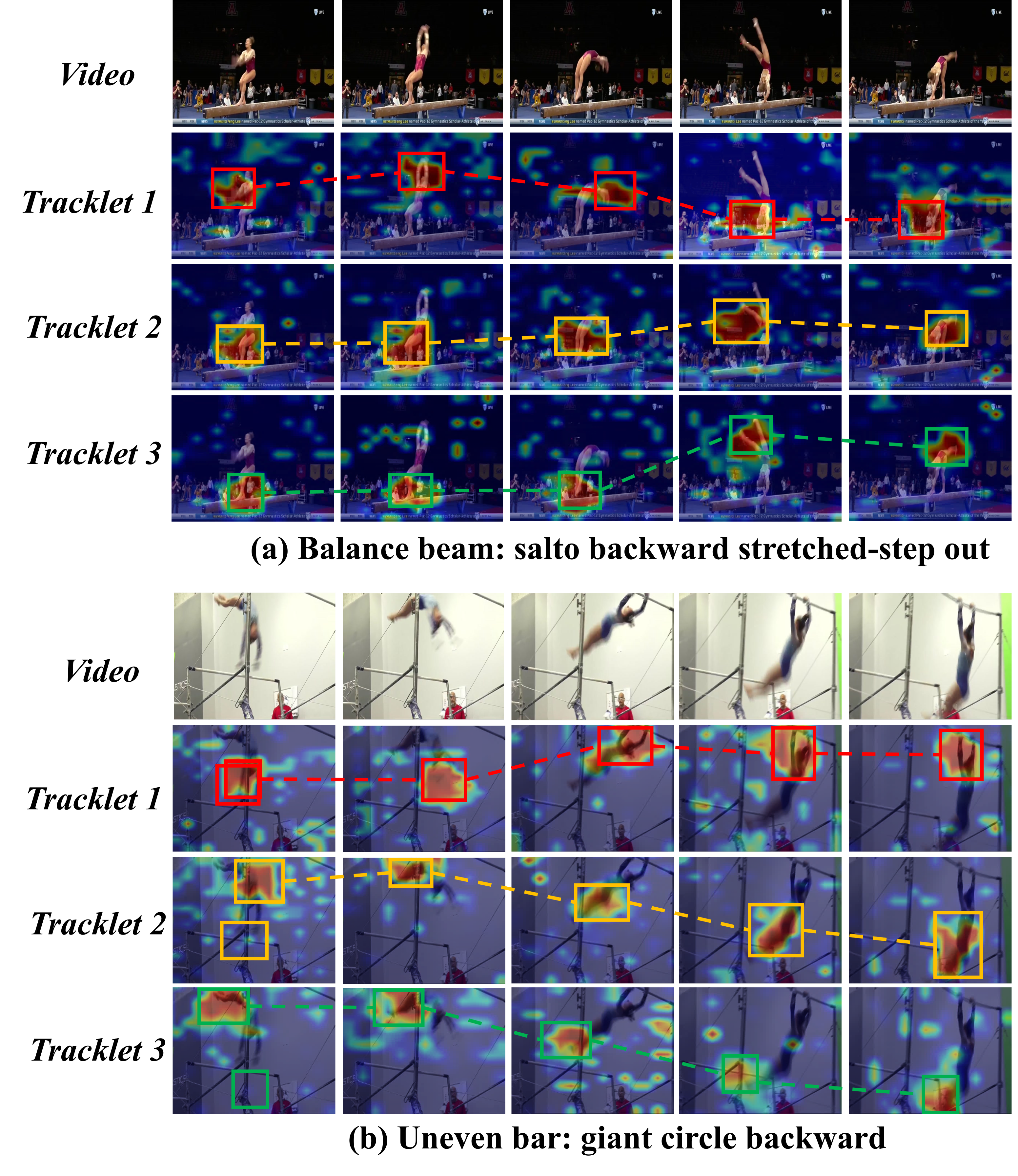}}
	\caption{\textbf{Qualitative visualization of three region-specific tracklets extracted by ART across two fine-grained gymnastics actions.} For each video, the top row shows original frames, while the bottom rows overlay region-specific responses to three semantic queries. Bounding boxes and dashed lines indicate spatial activations and their temporal coherence. ART reliably tracks fine-grained action parts—such as leading legs, gripping hands, and torsional joints—with clear spatio-temporal consistency.}
	\label{fig:vism2}
\end{figure}

\subsection{Complexity Analysis.}
To further assess the efficiency-performance trade-off of the proposed ART framework, we present a comparison of computational complexity, model size, and recognition accuracy against a strong baseline, UniFormerV2, as summarized in Table \ref{table:GFLOPs}. The ART model incurs a modest increase in GFLOPs, rising from 1332.43 to 1429.32, representing a 7.27\% increase in computational cost. Similarly, the parameter count grows from 345.05 million to 365.01 million, an increase of 5.78\%. Despite these slight increases in resource demand, ART achieves a substantial performance gain, improving the Top-1 accuracy from 90.5\% to 93.5\% (+3.0\%) and the Mean accuracy from 85.7\% to 89.2\% (+3.5\%). We conduct a training time comparison experiment using
four NVIDIA A40 GPUs with a batch size of 24 on the FineGym99 dataset.
The results are summarized in Table \ref{table:GFLOPs}. Despite the inclusion of six loss terms,
the epoch training time of ART is 121.8 minutes, only moderately higher than
UniFormerV2 (110.4 minutes).
Moreover, our ART framework achieves a competitive balance between accuracy and speed, with an average inference time of 36.2 ms per video clip, which is comparable to UniFormerV2 (35.8 ms), while maintaining superior accuracy.

These results highlight ART’s ability to achieve a more discriminative spatio-temporal representation with only a marginal increase in complexity. This trade-off is particularly favorable in contexts where fine-grained action recognition accuracy is critical, such as competitive sports analysis, surgical skill assessment, or human-computer interaction systems, where small improvements in accuracy can lead to significant downstream impact. Overall, the comparative analysis confirms that ART is not only effective but also computationally feasible for high-precision video understanding tasks.

\subsection{Visualization Analysis}
In Figure \ref{fig:vis}, we visualize the differences between the class activation maps (CAM) \cite{DBLP:conf/cvpr/ZhouKLOT16} generated by the baseline backbone network and the region-specific semantic response maps obtained from ART, across three representative action categories from the FineGym288 dataset. These include: (a) \textit{Switch leap (leap forward with leg change)}, (b) \textit{Split jump}, and (c) \textit{Flic-flac with step-out, also with support on one arm}.
In the first two examples (a) and (b), which involve well-defined and structured body motions, ART demonstrates a clear advantage in tracking temporally coherent, action-relevant regions. Unlike CAM, which typically highlights a coarse blob around the general human figure, ART produces spatially focused and temporally consistent activation over discriminative body parts such as the raised leg or extended thigh. This fine-grained attention effectively captures subtle variations in movement that are critical for distinguishing between visually similar action classes.
            
In complex high-velocity motions with asymmetric support, ART fails to consistently track the supporting arm, overemphasizing torso and legs, revealing limitations under rapid deformations or occlusions. Nonetheless, visualizations show that our tracklets generally capture relevant regions and maintain temporal consistency across frames.
To validate ART’s ability to capture fine-grained dynamics, Figure \ref{fig:vism2} shows three region tracklets over time. ART consistently tracks semantically meaningful parts—e.g., legs, torso, and arms in floor gymnastics, or hands, shoulders, and legs on uneven bars—demonstrating improved region localization, temporal alignment, and semantic consistency for modeling subtle action variations.


\section{Conclusion}
We present a novel discriminative actions tracking framework, namely ART, built upon transformers for fine-grained action recognition. Specifically, given a video consisting of multiple frames and a set of learnable vectors serving as the distinctive region query for each frame, ART firstly extracts region-specific semantic responses from video frames with distinctive region queries via the self-attention mechanism, and then forms a group of action tracklets for better characterizing fine-grained actions.
Moreover, to capture diverse semantic responses in each frame and model the correlation of similar response regions cross frames, we introduce a Multi-level Tracklet Contrastive Loss (MTC-Loss) among multiple region responses at spatial, temporal and tracklet levels. Comprehensive experimental results on four challenging dataests clearly demonstrate the superiority of our proposed ART.


%


\bibliographystyle{IEEEtran}
\bibliography{mybib}

\begin{IEEEbiography}[{\includegraphics[width=1in,height=1.25in,clip,keepaspectratio]{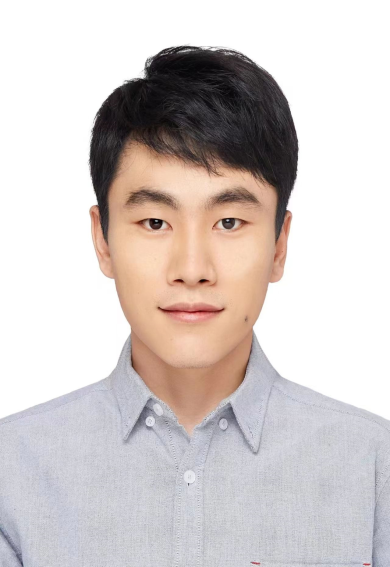}}]{Baoli Sun} is currently pursuing the Ph.D. degree with the School of Software Technology, Dalian University of Technology. He received the B.S degree in microelectronics science and engineering in 2018 from the Hefei University of Technology , Anhui , China.  He received his M.S. degree in software engineering in 2021 from the Dalian University of Technology, Dalian, China. His research interests include computer vision and video understanding.
\vspace{-2mm}
\end{IEEEbiography}

\begin{IEEEbiography}[{\includegraphics[width=1in,height=1.25in,clip,keepaspectratio]{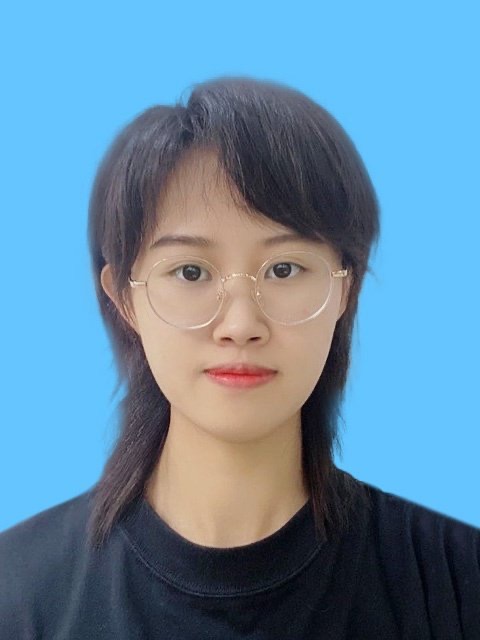}}]{Yihan Wang}
received the B.E. degree in software engineering in 2022 from the Dalian University of Technology, Dalian, China.  She is currently working toward the M.S. degree with the School of Software Technology, Dalian University of Technology. Her research interests include computer vision and video understanding.
\vspace{-2mm}
\end{IEEEbiography}

\begin{IEEEbiography}[{\includegraphics[width=1in,height=1.25in,clip,keepaspectratio]{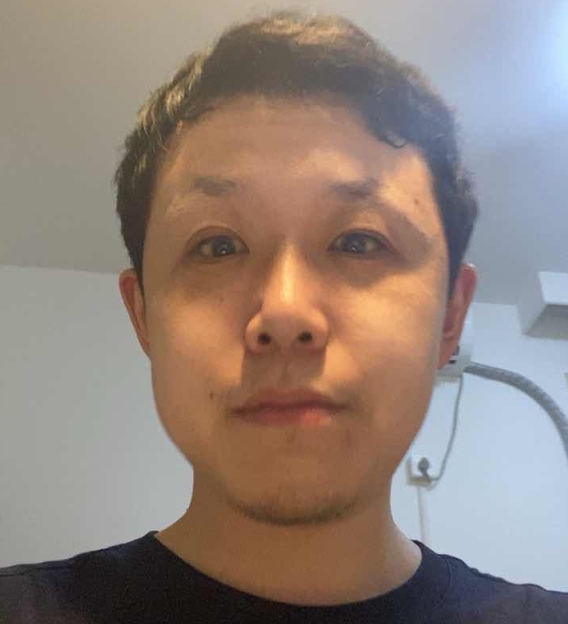}}]{Xinzhu Ma} received his B.Eng and M.P’s degree from Dalian University of Technology in 2017 and 2019, respectively. After that, He got the Ph.D degree from the University of Sydney in 2023. He is currently a postdoctoral researcher at the Chinese University of Hong Kong. His research interests include deep learning and computer vision.
\vspace{-2mm}
\end{IEEEbiography}

\begin{IEEEbiography}[{\includegraphics[width=1in,height=1.25in,clip,keepaspectratio]{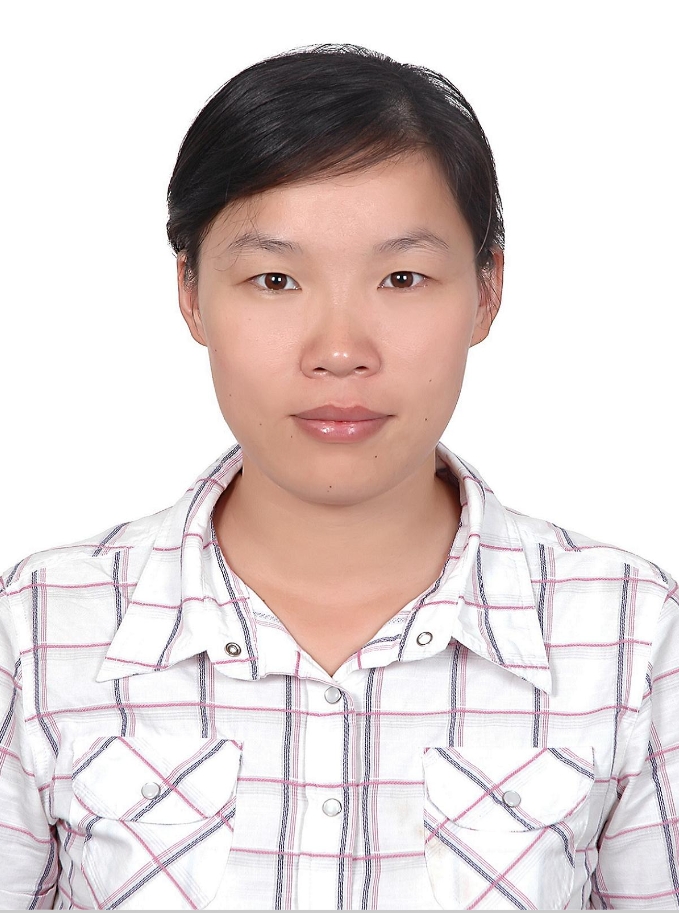}}]{Zhihui Wang}
received the B.S. degree in software engineering in 2004 from the North Eastern University, Shenyang, China. She received her M.S. degree in software engineering in 2007 and the Ph.D degree in software and theory of computer in 2010, both from the Dalian University of Technology, Dalian, China. Since November 2011, she has been a visiting scholar of University of Washington. Her current research interests include information hiding and image compression.
\vspace{-2mm}
\end{IEEEbiography}
\begin{IEEEbiography}[{\includegraphics[width=1in,height=1.25in,clip,keepaspectratio]{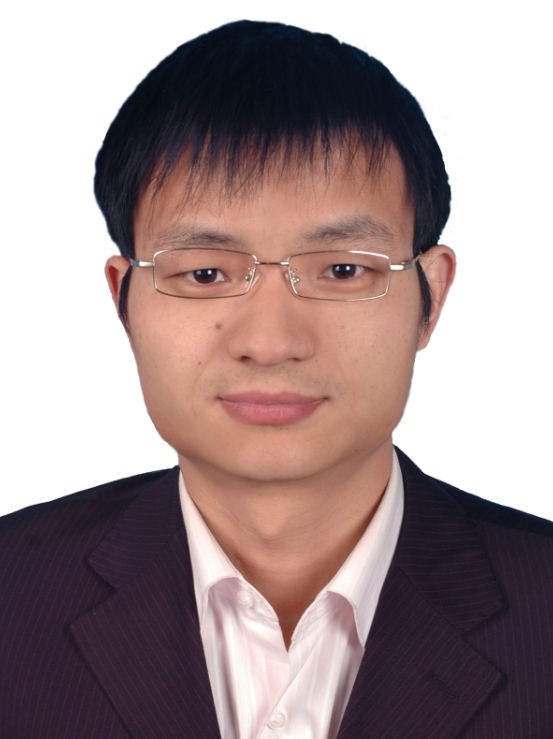}}]{Kun Lu}
    received the B.S. degree in computer science and engineering in 2002 from the Dalian University of Technology, Dalian, China. He received her M.S. degree in software engineering in 2005 and the Ph.D degree in software and theory of computer in 2017, both from the Dalian University of Technology, Dalian, China. 
 His current research interests include mobile swarm sensing, machine learning, edge computing and big data analytics.
    \vspace{-2mm}
    \end{IEEEbiography}

\begin{IEEEbiography}[{\includegraphics[width=1in,height=1.25in,clip,keepaspectratio]{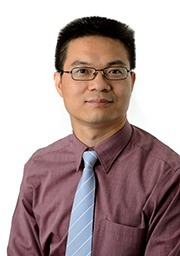}}]{Zhiyong Wang}
received his B.Eng. and M.Eng. degrees in electronic engineering from South China University of Technology, Guangzhou, China, and his Ph.D. degree from Hong Kong Polytechnic University, Hong Kong. He is a Senior Lecturer of the School of Information Technologies, the University of Sydney, after joining the school as a Postdoctoral Research Fellow in 2003. His research interests include multimedia information processing, retrieval and management, human-centered multimedia computing, pattern recognition, and machine learning.
\vspace{-2mm}
\end{IEEEbiography}

\end{document}